\documentclass{article}

\PassOptionsToPackage{numbers, sort&compress}{natbib}
\usepackage[preprint]{neurips_2026}

\usepackage[utf8]{inputenc} % allow utf-8 input
\usepackage[T1]{fontenc}    % use 8-bit T1 fonts
\usepackage{hyperref}       % hyperlinks
\usepackage{url}            % simple URL typesetting
\usepackage{booktabs}       % professional-quality tables
\usepackage{multirow}       % multi-row cells in tables
\usepackage{amsfonts}       % blackboard math symbols
\usepackage{nicefrac}       % compact symbols for 1/2, etc.
\usepackage{microtype}      % microtypography
\usepackage{xcolor}         % colors
\usepackage{graphicx}       % includegraphics
\usepackage{xspace}
\usepackage{algorithm}
\usepackage{algpseudocode}
\usepackage{amsmath}

\makeatletter
\DeclareRobustCommand\onedot{\futurelet\@let@token\@onedot}
\def\@onedot{\ifx\@let@token.\else.\null\fi\xspace}

\def\eg{\emph{e.g}\onedot} 

\def\ie{\emph{i.e}\onedot}

\makeatother

\def\method{X-WAM\xspace}

\title{Unified 4D World Action Modeling from Video Priors with Asynchronous Denoising}

\author{%
  Jun Guo$^{1,2}$
  \quad
  Qiwei Li$^{2,3}$
  \quad
  Peiyan Li$^{2,4}$
  \quad
  Zilong Chen$^{1}$
  \quad
  Nan Sun$^{1,2}$
  \\
  \textbf{Yifei Su}$^{2}$
  \quad
  \textbf{Heyun Wang}$^{2}$
  \quad
  \textbf{Yuan Zhang}$^{2}$
  \quad
  \textbf{Xinghang Li}$^{2\dagger}$
  \quad
  \textbf{Huaping Liu}$^{1\dagger}$
  \\
  \\
  $^1$Tsinghua University \quad $^2$Xiaomi Robotics \quad
  $^3$ Peking University \quad $^4$ CASIA
  \\
  \\
  \url{https://sharinka0715.github.io/X-WAM/}
}

\begin{document}

\maketitle

\begin{abstract}
  % Current embodied AI models are fragmented into separate paradigms: policy models (VLAs, WAMs) that focus on action prediction for robot control, and world models that focus on simulating future observations. Each addresses a single objective with inherent limitations. Recent efforts have begun unifying video generation and action prediction within a single framework, demonstrating the potential of multi-task joint modeling. However, these unified approaches remain confined to 2D pixel-space modeling, lacking explicit spatial awareness. In this work, we take a further step by incorporating spatial information into the unified modeling paradigm. 
  We propose \method, a Unified 4D World Model that unifies real-time robotic action execution and high-fidelity 4D world synthesis (video + 3D reconstruction) in a single framework, addressing the critical limitations of prior unified world models (\textit{e.g.}, UWM) that only model 2D pixel-space and fail to balance action efficiency and world modeling quality. To leverage the strong visual priors of pretrained video diffusion models, \method imagines the future world by predicting multi-view RGB-D videos, and obtains spatial information efficiently through a lightweight structural adaptation: replicating the final few blocks of the pretrained Diffusion Transformer into a dedicated depth prediction branch for the reconstruction of future spatial information. Moreover, we propose Asynchronous Noise Sampling (ANS) to jointly optimize generation quality and action decoding efficiency. ANS applies a specialized asynchronous denoising schedule during inference, which rapidly decodes actions with fewer steps to enable efficient real-time execution, while dedicating the full sequence of steps to generate high-fidelity video. Rather than entirely decoupling the timesteps during training, ANS samples from their joint distribution to align with the inference distribution. Pretrained on over 5,800 hours of robotic data, \method achieves 79.2\% and 90.7\% average success rate on RoboCasa and RoboTwin 2.0 benchmarks, while producing high-fidelity 4D reconstruction and generation surpassing existing methods in both visual and geometric metrics.
\end{abstract}

\section{Introduction}

The pursuit of general-purpose Embodied AI has been significantly accelerated by the advent of robotic foundation models. Current approaches in this space can be broadly categorized into two paradigms, each targeting a single objective. On the one hand, \emph{policy models} focus on predicting executable actions for robot control. Vision-Language-Action (VLA) models~\cite{rt2, octo, openvla, pi0, pi0.5, gr00t, gr1} fine-tune pretrained Vision-Language Models (VLMs) to output motor commands, excelling at instruction following and semantic reasoning but lacking the geometric intuition and physical awareness of how actions continuously unfold in the real world~\cite{lingbotva}. World Action Models (WAMs)~\cite{lingbotva, dreamzero, fastwam, cosmospolicy, gigaworldpolicy} further leverage video generation models to jointly predict future observations and actions, harnessing video priors for stronger physical understanding and generalization. On the other hand, World Models~\cite{dreamerv3, irasim, genie, cosmos, gigaworld, emu35} focus on simulating future observations: text-conditioned and action-conditioned world models excel at generating realistic visual predictions of physical dynamics, but do not directly produce executable actions for robot control. These separate paradigms each address a single task, limiting cross-task synergy and representational efficiency.

Recently, a line of work has begun to bridge this divide by constructing unified world action models~\cite{uwm, motus, videovla, genieenvisioner} that jointly model video generation and action prediction within a single framework. By sharing representations across modalities, these approaches achieve encouraging results in both future prediction quality and policy execution, demonstrating the significant potential of multi-task unified modeling. However, they remain confined to 2D pixel-space observation, lacking explicit spatial awareness and 3D geometric grounding. Since the physical world is fundamentally three-dimensional, this confinement strips away critical geometric structures, causing models to hallucinate physically implausible futures and preventing geometrically faithful 3D reconstruction. To unlock the full potential of unified world action models, it is imperative to elevate them from 2D pixel predictors to spatially aware 4D dynamics simulators that jointly address generation, reconstruction, and policy execution.

\begin{figure}[t]
\centering
\includegraphics[width=\textwidth]{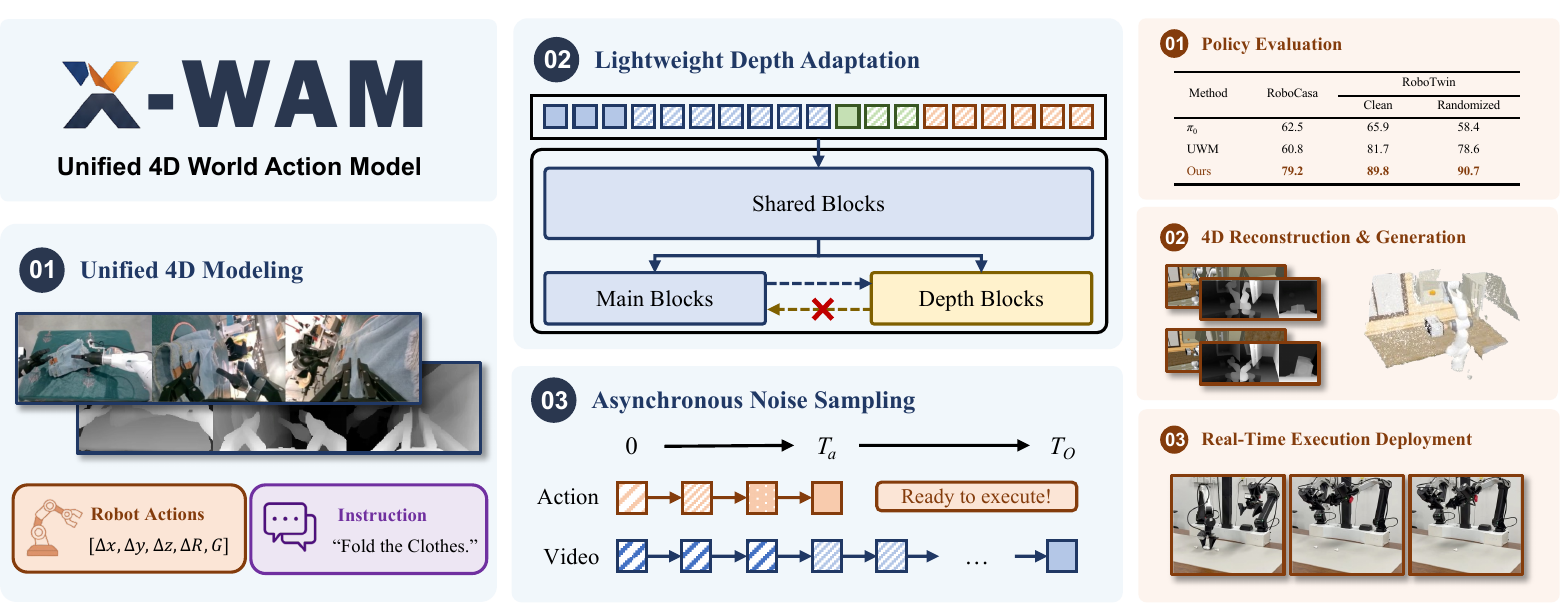}
\vspace{-0.1in}
\caption{\textbf{Overview of \method.} \textit{Top:} \method is a unified 4D World Action Model that jointly predicts future multi-view RGB-D videos and robot actions from video priors, featuring a lightweight depth adaptation module for spatial reconstruction and Asynchronous Noise Sampling (ANS) for efficient action decoding. \textit{Bottom:} \method surpasses existing methods in policy success rate on RoboCasa and RoboTwin 2.0, produces high-fidelity 4D reconstruction and generation, and enables real-time execution deployment on physical robots.}
\label{fig:teaser}
% \vspace{-0.2in}
\end{figure}

Building upon these initial unification efforts, we take a further step by incorporating explicit spatial information into the unified modeling paradigm. We propose \method (Figure~\ref{fig:teaser}), a unified 4D World Action Model that simultaneously targets four objectives within a single architecture: high-fidelity video generation, 3D spatial reconstruction, high policy success rate, and efficient action execution. Built on the powerful visual priors of the pretrained video foundation model, \method takes multi-view RGB observations and current robot states as inputs to jointly generate future 4D observations alongside the robot's future states and actions. However, seamlessly integrating 4D spatial awareness and policy execution into such a unified framework presents two fundamental technical challenges.

The first challenge lies in effectively injecting 3D perception into the model without destroying its pretrained knowledge or introducing prohibitive computational overhead. Naive approaches to spatial modeling face distinct drawbacks: concatenating depth maps as additional tokens along the sequence dimension doubles the sequence length, incurring quadratic growth in attention cost; alternatively, fusing depth along the channel dimension shifts the input distribution away from the pretrained manifold, substantially increasing learning difficulty. To circumvent these bottlenecks, we introduce a lightweight structural adaptation. Rather than expanding the denoising sequence, \method explicitly models the 4D world by simply replicating the final few blocks of the pretrained Diffusion Transformer (DiT)~\cite{dit} to construct a dedicated depth prediction branch. This elegant design successfully extracts 3D spatial information without altering the original model's core structure, bypassing the sequence length explosion while strictly preserving the integrity of the pretrained visual priors. As shown in our experiments, this depth supervision not only enables high-quality 3D reconstruction but also consistently improves policy success rates, confirming that explicit spatial modeling benefits multiple objectives of the unified framework simultaneously.

The second challenge stems from the inherent modality mismatch when jointly generating high-dimensional video trajectories and low-dimensional robotic actions. While synthesizing high-fidelity video necessitates numerous denoising steps~\cite{ddim, edm}, low-dimensional actions require far fewer steps~\cite{diffusionpolicy, pi0}, and can be accurately recovered even from highly noisy video latents~\cite{motus, mimicvideo, fastwam, dreamzero, dit4dit}. Motivated by this insight, we propose Asynchronous Noise Sampling (ANS). ANS introduces a specialized asynchronous denoising schedule for inference: it rapidly decodes precise actions using only a fraction of the initial steps to allow them to be immediately executed by the policy, and subsequently completes the remaining steps to render high-fidelity future videos. Driven by this asynchronous inference characteristic, ANS accordingly reformulates the training-stage sampling strategy. Instead of completely decoupling the noise timesteps of these modalities via independent random sampling, ANS systematically samples from a joint distribution of video and actions that strictly matches the test-time distribution. This elegantly eliminates the inefficiencies of decoupled sampling, maximizing both action inference speed and visual generation quality.

In summary, our primary contributions are threefold:
\begin{itemize}
    \item We propose \method, a unified 4D World Action Model that incorporates explicit 3D spatial awareness into the joint video-action modeling paradigm. By introducing a lightweight structural adaptation, replicating the final blocks of the pretrained DiT as a dedicated depth branch, we achieve high-quality spatial modeling without doubling sequence lengths or disrupting pretrained visual priors.
    \item We introduce Asynchronous Noise Sampling (ANS) to enhance the joint generation of videos and actions. By sampling from their joint distribution and employing an asynchronous denoising schedule, ANS improves the training efficiency while maximizing both action decoding speed and video generation quality.
    \item We demonstrate that \method consistently outperforms all baselines on RoboCasa and RoboTwin 2.0 benchmarks and real-world earphone packing experiments, while producing superior 4D reconstruction and generation across both visual and geometric metrics, validating that a single unified framework can jointly optimize policy execution, visual generation, and spatial reconstruction.
\end{itemize}

\section{Related Work}

\subsection{Unified World Action Modeling}

Current general embodied models fall into two complementary paradigms. \emph{Policy models}, predominantly Vision-Language-Action (VLA) models~\cite{rt2, octo, roboflamingo, openvla, pi0, pi0.5, gr00t, gr1, rdt, xr0}, map observations directly to executable robot actions for real-time control. \emph{World models}~\cite{dreamerv3, irasim, genie, cosmos, gigaworld, emu35} aim to model environmental dynamics and learn to imagine future observations. Although naturally complementary, the two paradigms have largely evolved in isolation. Some works bridge the gap from the world model side by attaching inverse dynamics models or extracting intermediate representations to convert world models into planners~\cite{unipi, vpp, genieenvisioner, vista, pi0.7}. Others augment VLAs with auxiliary future prediction objectives to inject dynamics awareness~\cite{moto, worldvla, dreamvla, univlahang, vlajepa, bagelvla}. While both directions yield improvements, they remain loosely coupled rather than truly unified.

Recently, a line of work has sought to build end-to-end World Action Models (WAMs) from video foundation models. UWM~\cite{uwm} and Motus~\cite{motus} formulate the problem as a Unified World Model, enabling flexible conditioning and multi-task generation. VideoVLA~\cite{videovla} and Cosmos Policy~\cite{cosmospolicy} directly append action tokens into video sequences for joint prediction. Other works~\cite{mimicvideo, dit4dit, fastwam} employ a Mixture of Transformer architecture with independent parameters and denoising timesteps for each modality. DreamZero~\cite{dreamzero}, LingBot-VA~\cite{lingbotva}, and GigaWorld-Policy~\cite{gigaworldpolicy} leverage causal attention masks and KV caching to reduce inference latency. Surveys~\cite{wamsurvey} have shown that such unified approaches generalize better than traditional VLAs. Despite this progress, two limitations persist. First, existing unified models remain confined to 2D pixel-space, lacking explicit 3D spatial awareness. Second, although several recent WAMs~\cite{motus, uwm, dreamzero} have begun to address the balance between video generation quality and action decoding efficiency by decoupling their sampling timesteps, these approaches rely on independent noise sampling for each modality during training, which introduces a mismatch between the training and inference distributions and thus limits the effectiveness of asynchronous denoising.

\subsection{3D Modeling in Embodied Models}

Owing to the abundance and accessibility of 2D data, contemporary mainstream embodied models primarily operate within a 2D space for perception, modeling, and prediction. However, the lack of explicit spatial awareness and modeling capabilities, coupled with an over-reliance on purely data-driven fitting, creates a significant bottleneck in tasks that demand spatial comprehension and out-of-distribution generalization. To address this issue, numerous studies have incorporated 3D information into the training pipeline to further enhance the capabilities of embodied models. Within the VLA framework, one category of research~\cite{3dvla, dreamvla, spatial-vla, evo0, spatialforcing, falcon} encodes 3D features to serve as predictive targets or supervisory signals within the model's sequence. Another category~\cite{pointvla, geovla, bridgevla} directly utilizes explicit 3D representations as inputs, performing predictions natively within the 3D space.

In the context of world models and world action models, several approaches~\cite{tesseract, flowdreamer, enerverse, robot4dgen, pointworld, wristworld, mvista4d} introduce 3D supervisory signals during the video generation process, endowing the models with multi-view consistency and superior spatial reasoning. ManiGaussian~\cite{manigaussian} and GWM~\cite{gwm} construct world models entirely within 3D representations, utilizing the neural rendering technique of 3D Gaussian Splatting~\cite{3dgs} to build high-fidelity 3D world models. Given that current open-source robotic datasets are predominantly composed of 2D videos, existing 3D modeling methods frequently rely on pre-trained feed-forward 3D reconstruction models~\cite{dust3r, vggt, vda, da3} to extract spatial information from robotic data. This strategy effectively transfers the spatial reasoning capabilities of reconstruction models to the video generation pipelines. To the best of our knowledge, no existing work has incorporated explicit spatial information into the unified world action modeling paradigm, nor has any unified model demonstrated the ability to simultaneously serve as a high-fidelity video generator, a 3D reconstruction system, and an efficient policy model within a single framework.

\section{Methodology}

\begin{figure}[t]
\centering
\includegraphics[width=\textwidth]{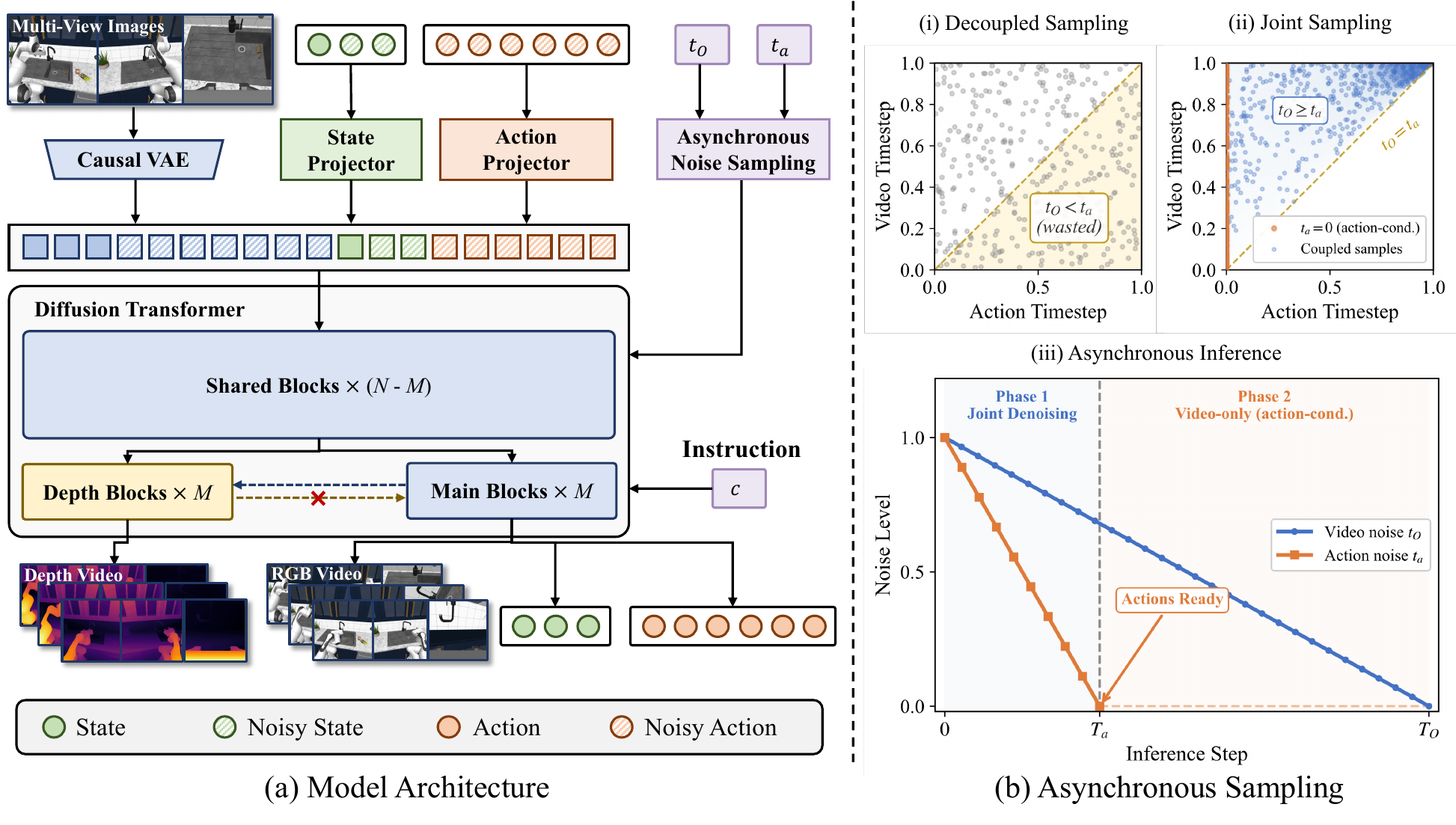}
\vspace{-0.1in}
\caption{\textbf{Overview of \method.} (a) Model architecture: multi-view RGB observations, proprioceptive states, and noisy actions are encoded and jointly denoised by a Diffusion Transformer initialized from Wan2.2-5B, with a lightweight interleaved depth branch for spatial modeling. (b) Asynchronous Noise Sampling (ANS): (i) standard decoupled sampling wastes training on configurations where $t_O < t_a$; (ii) our coupled joint sampling ensures $t_O \geq t_a$, faithfully matching the inference distribution; (iii) during inference, actions are decoded in $T_a$ steps and immediately dispatched, while video denoising continues for $T_O$ steps.}
\label{fig:framework}
% \vspace{-0.2in}
\end{figure}

Building upon recent advances in unified world action modeling, we propose \method as a unified framework that simultaneously addresses video generation, 3D spatial reconstruction, policy success rate, and efficient action execution. This is achieved through two core designs: a lightweight depth adaptation module (Section~\ref{sec:depth_adapt}) that enables spatial reconstruction, and Asynchronous Noise Sampling (Section~\ref{sec:ans}) that jointly optimizes generation quality and action decoding efficiency. We begin by presenting the overall model architecture (Section~\ref{sec:architecture}), and then detail the training procedure including data processing and the training pipeline (Section~\ref{sec:training}).

\subsection{Model Architecture}
\label{sec:architecture}

\method takes a language instruction $c$, the initial proprioceptive state $s_0$, and multi-view initial RGB observations $O_0$ as conditions, and jointly predicts future RGB videos $O_{1:H}$, depth videos $D_{1:H}$, proprioceptive states $s_{1:H}$, and actions $a_{1:K}$, where $H$ and $K$ denote the prediction horizons for video/state and actions, respectively. Following~\cite{videovla, dreamzero}, \method is fine-tuned from a pretrained video generation Diffusion Transformer~\cite{dit}, specifically Wan2.2-TI2V-5B~\cite{wan} in this work. RGB videos are encoded into latent representations via the original causal VAE encoder $\mathcal{E}$, \ie, $\mathbf{z}_{O} = \mathcal{E}(O)$, while proprioceptive states and robot actions are projected into the latent space via learnable MLPs: $\mathbf{z}_{s} = \mathrm{MLP}_s(s)$ and $\mathbf{z}_{a} = \mathrm{MLP}_a(a)$. After denoising, the clean latents are decoded back to the original physical spaces via symmetric MLPs: $\hat{s} = \mathrm{MLP}_s^{\text{dec}}(\mathbf{z}_s)$ and $\hat{a} = \mathrm{MLP}_a^{\text{dec}}(\mathbf{z}_a)$. The three modalities are concatenated into a unified denoising sequence:
\begin{equation}
    \mathbf{Z} = [\mathbf{z}_{O_0},\, \mathbf{z}_{O_{1:H}},\, \mathbf{z}_{s_0},\, \mathbf{z}_{s_{1:H}},\, \mathbf{z}_{a_{1:K}}],
\end{equation}
which is processed with bidirectional full attention, with depth reconstructed from the generated RGB video sequence. The initial observation $\mathbf{z}_{O_0}$ and state $\mathbf{z}_{s_0}$ remain fixed throughout the denoising process with their noise timestep set to $t=0$ (\ie, treated as clean samples).

Concretely, given 1 conditioning RGB frame and 1 initial state, \method predicts $H=8$ future RGB frames, $H=8$ future states, and $K=32$ future actions. This asymmetric design reflects the different temporal requirements of each modality: actions demand a higher control frequency for smooth and responsive robot execution, while RGB frames and states can be predicted at a lower frequency sufficient for visual generation and 4D reconstruction. The states are temporally aligned with the video frames to enable frame-wise multi-view RGB-D fusion for 3D reconstruction, whereas the actions are uniformly distributed across the same time span at $K/H = 4\times$ the video frame rate.

The original video diffusion model is designed for single-view 2D generation, employing 3D Rotary Position Embeddings (RoPE)~\cite{rope} to encode temporal and spatial positions within the sequence. To enable multi-view compatibility without disrupting the pretrained positional encodings, we augment the tokens of each viewpoint with learnable view embeddings to indicate the view index. For proprioceptive states and actions, we apply the same temporal RoPE as the video tokens along the time dimension, allowing the model to infer the temporal correspondence between states/actions and video frames through positional proximity.

\method is designed to simultaneously reconstruct and generate the future world. Reconstructing 3D representations (\eg, point clouds) from multi-view RGB-D outputs requires camera poses for each viewpoint. Unlike prior works~\cite{enerverse} that explicitly encode camera extrinsics or ray direction maps as tokens, we adopt a more principled approach grounded in the structure of robotic systems. We observe that cameras in robotic manipulation setups can be categorized into two types: \emph{static} cameras (first-person and third-person views), whose poses remain constant throughout task execution, and \emph{dynamic} cameras (wrist-mounted), which are rigidly attached to the robot arm with a fixed hand-eye calibration matrix that depends solely on the robot model. Therefore, instead of predicting explicit camera extrinsics, \method predicts the end-effector pose $\mathbf{T}_{\text{ee}} \in SE(3)$ and derives the wrist camera pose via the fixed hand-to-eye calibration matrix $\mathbf{T}_{\text{h2e}}$:
\begin{equation}
    \mathbf{T}_{\text{wrist}} = \mathbf{T}_{\text{ee}} \cdot \mathbf{T}_{\text{h2e}}.
\end{equation}
This conversion enables the fusion of 3D information across all viewpoints to reconstruct a unified 3D representation.

\subsection{Lightweight Depth Adaptation}
\label{sec:depth_adapt}

In \method, depth maps are predicted in the same VAE latent space as RGB frames: each single-channel depth map is replicated three times along the channel dimension to form a pseudo-RGB image, encoded by the same causal VAE $\mathcal{E}$, and decoded back via the symmetric VAE decoder after denoising. Rather than concatenating depth tokens along the sequence dimension (which doubles attention cost) or fusing along the channel dimension (which shifts the input distribution away from the pretrained manifold), we propose a lightweight depth adaptation module that modifies the pretrained DiT architecture. Specifically, given a model with $N$ DiT blocks, we replicate the final $M$ blocks ($M < N$) to construct an auxiliary depth prediction branch. After the shared first $N\!-\!M$ blocks produce hidden states $\mathbf{H}$, the depth branch and the main branch are initialized as $\mathbf{Z}_D^{(0)} = \mathbf{Z}_{\text{m}}^{(0)} = \mathbf{H}$ and executed in an \emph{interleaved} fashion. At each layer $j \in \{1, \dots, M\}$:
\begin{equation}
    \mathbf{Z}_D^{(j)} = \mathrm{DepthBlock}_j\!\left(\mathbf{Z}_D^{(j-1)} \mid \mathbf{Z}_{\text{m}}^{(j-1)}\right), \quad
    \mathbf{Z}_{\text{m}}^{(j)} = \mathrm{DiTBlock}_{N-M+j}\!\left(\mathbf{Z}_{\text{m}}^{(j-1)}\right),
\end{equation}
where $\mathrm{DepthBlock}_j$ attends to the main branch's input $\mathbf{Z}_{\text{m}}^{(j-1)}$ at the same layer via cross-attention, while the main branch remains unaffected by depth tokens. We term this asymmetric connectivity \emph{unilateral attention}: the depth branch can read from the main branch, but not vice versa, thereby strictly preserving the integrity of the pretrained weights. The depth branch is trained to regress the inverse depth of the current video frame using mean squared error (MSE) loss, consistent with established depth estimation models~\cite{vda, da3}.

This design rests on the fundamental assumption that depth information can be inferred from RGB features without requiring fully independent generation. The detailed single-step procedure is presented in Algorithm~\ref{alg:single_step} (Appendix~\ref{app:algorithms}). Depth supervision during training enhances the model's spatial structure perception. During inference, this auxiliary depth branch does not need to participate in every denoising step and can be flexibly toggled on or off, substantially reducing rollout overhead.

\subsection{Asynchronous Noise Sampling}
\label{sec:ans}

Prior works~\cite{motus, mimicvideo, dreamzero, dit4dit, fastwam} decouple the noise timesteps of video and actions to allow faster action decoding during inference. However, independently sampling each modality's timestep introduces training configurations that never arise at test time (\eg, $t_O < t_a$), degrading training efficiency. To align the training and inference noise distributions, \method adopts Asynchronous Noise Sampling (ANS). The complete procedure is detailed in Algorithm~\ref{alg:ans} (Appendix~\ref{app:algorithms}).

\paragraph{Asynchronous inference.} During inference, ANS applies asynchronous denoising timesteps for video and actions, as illustrated in Figure~\ref{fig:framework}(b-iii). We allocate $T_a$ denoising steps for proprioceptive states and actions, and $T_O$ denoising steps for video ($T_a < T_O$). Both modalities start from pure noise and are denoised with step sizes of $\frac{1}{T_a}$ and $\frac{1}{T_O}$, respectively. After $T_a$ forward passes, the noise-free actions are obtained and can be immediately dispatched to the downstream robot for execution. If a clear, complete video is desired, the remaining $T_O - T_a$ steps are continued, during which the actions serve as a clean modality and undergo no further denoising. In this regime, the inference process naturally becomes an action-conditioned world model.

\paragraph{Coupled noise sampling during training.} To realize this asynchronous inference behavior, we also apply different noise timesteps to video and actions during training. However, unlike prior works that sample from two independent distributions, we place the video and action noise levels into a joint distribution and perform coupled sampling. Formally, the joint noise level $(t_O, t_a)$ is drawn from the following mixture:
\begin{equation}
    (t_O, t_a) \sim
    \begin{cases}
        t_a = 0,\;\; t_O \sim \mathrm{U}(0, 1) & \text{w.p. } p, \\[4pt]
        t_a \sim \mathrm{U}(0, 1),\;\; t_O = t_a + (1 - t_a) \cdot b,\;\; b \sim \mathrm{Beta}(1.5, 1) & \text{w.p. } 1\!-\!p,
    \end{cases}
    \label{eq:ans_joint}
\end{equation}
where the first case corresponds to action-conditioned video generation with noise-free actions, and the second case represents asynchronous joint generation. The $\mathrm{Beta}(1.5, 1)$ distribution, rescaled to $[t_a, 1]$, biases $t_O$ toward higher noise levels, reflecting the fact that video typically requires more denoising steps than actions. Crucially, $t_O$ is sampled \emph{conditioned on} $t_a$, making them dependent rather than independent random variables. This coupled sampling strategy more faithfully reflects the inference-time distribution, enabling more efficient training of the WAM.

\subsection{Training Details}
\label{sec:training}

Consistent with the pretrained Wan2.2-5B model~\cite{wan}, we fine-tune \method using the flow matching framework~\cite{flowmatching}. The model $f_\theta$ is trained to predict the velocity field $\mathbf{v} = \boldsymbol{\epsilon} - \mathbf{z}^0$ given noisy inputs at timestep $t$. For a modality $m \in \{O, s, a\}$ with corresponding timestep $t_m$, the velocity prediction loss is:
\begin{equation}
    \mathcal{L}_{m} = \left\| f_\theta^{m}(\mathbf{z}_m^{t_m}, t_m) - (\boldsymbol{\epsilon}_m - \mathbf{z}_m^0) \right\|^2,
\end{equation}
where $t_O$ and $t_a$ denote the video and action noise timesteps sampled via ANS (Eq.~\ref{eq:ans_joint}), with $t_s = t_a$. The depth branch is supervised with a direct MSE regression loss on inverse depth: $\mathcal{L}_{\text{depth}} = \left\| \hat{D} - D^{*} \right\|^2$, where $D^{*}$ denotes the ground-truth inverse depth. The total training objective is:
\begin{equation}
    \mathcal{L}_{\text{total}} = \mathcal{L}_{O} + \lambda_s \mathcal{L}_{s} + \lambda_a \mathcal{L}_{a} + \lambda_D \mathcal{L}_{\text{depth}},
\end{equation}
where $\lambda_s$, $\lambda_a$, and $\lambda_D$ are weighting coefficients.

To build a unified 4D model capable of generation, reconstruction, and manipulation, we train the model on over 5{,}800 hours of data, encompassing both real-robot and simulated datasets spanning diverse manipulation tasks. All datasets undergo preprocessing and filtering, and are unified into a consistent coordinate system and representation. We define a universal interface to represent robot states and actions across heterogeneous datasets. The state is defined as the poses and gripper positions of a dual-arm end-effector, while the action is defined as the corresponding changes in end-effector poses and gripper positions. For single-arm robots, we treat the single arm as the left arm and do not supervise the right-arm output. During inference, we employ the UniPC~\cite{unipc} multistep scheduler recommended by Wan2.2, maintaining separate scheduler instances with different step sizes for video and state/action modalities, following the asynchronous inference procedure described in Section~\ref{sec:ans}. Full details on pretraining data, implementation, and baselines are provided in Appendix~\ref{app:training_details}.
\section{Experiments}

We evaluate \method across three complementary dimensions: policy execution (Section~\ref{sec:exp_policy}), 4D reconstruction and generation (Section~\ref{sec:exp_4d}), and ablation studies that jointly analyze both objectives (Section~\ref{sec:exp_ablation}). We additionally deploy \method on a real-world dual-arm platform for earphone packing tasks (Appendix~\ref{app:real_robot}). This comprehensive evaluation validates that \method's unified framework simultaneously achieves strong performance across all dimensions.

\subsection{Policy Evaluation}
\label{sec:exp_policy}

We first evaluate the policy execution capability of \method by deploying it in closed-loop simulation and measuring task success rates on two representative robotic manipulation benchmarks.

\textbf{RoboCasa}~\cite{robocasa} is a large-scale simulation benchmark featuring diverse kitchen manipulation tasks with realistic scenes and object variations. We report the average success rate (SR) across 24 manipulation tasks. We compare against two VLA baselines: $\pi_0$~\cite{pi0} and GR00T-N1.5~\cite{gr00t}, and three WAM baselines: UWM~\cite{uwm}, DreamZero~\cite{dreamzero}, and Cosmos Policy~\cite{cosmospolicy}.

\textbf{RoboTwin 2.0}~\cite{robotwin2} is a dual-arm manipulation benchmark that evaluates policy generalization under two settings: \emph{Clean}, where the environment matches a certain distribution, and \emph{Randomized}, where object poses, appearances, and distractors are randomized to test robustness. Following~\cite{motus}, we train \method on all trajectories of AgileX arms, including 50 clean and 500 randomized trajectories on 50 tasks. We compare against two VLA baselines: $\pi_0$~\cite{pi0} and $\pi_{0.5}$~\cite{pi0.5}, and three WAM baselines: UWM~\cite{uwm}, GigaWorld-Policy~\cite{gigaworldpolicy}, and Motus~\cite{motus}.

Results on RoboCasa and RoboTwin 2.0 are presented in Table~\ref{tab:robocasa} and Table~\ref{tab:robotwin}, respectively. As shown in Table~\ref{tab:robocasa} and Table~\ref{tab:robotwin}, \method consistently outperforms all baselines on both benchmarks. On RoboCasa, \method attains 79.2\% average SR, surpassing the strongest baseline Cosmos Policy (67.1\%) by 12.1 percentage points. On RoboTwin 2.0, \method achieves 89.8\% and 90.7\% under the Clean and Randomized settings, respectively, outperforming the prior method Motus (88.7\% / 87.0\%) across both protocols. These results validate that incorporating explicit 3D spatial awareness and large-scale pretraining into the unified world action modeling framework yields substantial performance gains. Per-task breakdowns are provided in Appendix~\ref{app:detailed_results}.

\subsection{4D Reconstruction and Generation}
\label{sec:exp_4d}

We next evaluate the 4D reconstruction and generation capabilities of \method on the RoboCasa environment. Specifically, we execute the policy in simulation and compare the predicted multi-view RGB-D observations against the ground-truth observations rendered by the simulator. We adopt three groups of metrics: PSNR, SSIM, and LPIPS for visual fidelity, absolute relative error (AbsRel) and $\delta_1$ accuracy for depth quality, and Chamfer Distance (CD) for the quality of the reconstructed point clouds. The depth metrics (AbsRel, $\delta_1$) evaluate single-view reconstruction capability, measuring how accurately the model predicts per-pixel geometry from each individual viewpoint. In contrast, Chamfer Distance evaluates multi-view consistency and the accuracy of camera extrinsic estimation (for static cameras) and robot proprioceptive state prediction (for wrist cameras), as it requires lifting per-view depth maps into a shared 3D coordinate frame and fusing them into a unified point cloud. Note that pixel-level metrics (PSNR, SSIM, LPIPS, AbsRel, $\delta_1$) are computed only on the two static cameras (first-person and third-person views), as the wrist camera suffers from pixel misalignment due to minor errors in the predicted end-effector pose, rendering per-pixel comparison unreliable. The quality of wrist-camera predictions can instead be assessed through Chamfer Distance and qualitative visualizations. As baselines, we consider a two-stage approach that combines DreamZero~\cite{dreamzero} for RGB video generation with Depth Anything 3~\cite{da3} for post-hoc depth estimation, and Robot4DGen~\cite{robot4dgen}, a geometry-aware 4D video generation method. We also include an ablative variant, \method w/o depth + DA3, which removes our depth branch and instead relies on Depth Anything 3 for depth estimation.

\begin{table}[t]
\centering
\begin{minipage}[t]{0.48\textwidth}
\centering
\caption{Average success rate (\%) on 24 manipulation tasks of RoboCasa benchmark.}
\label{tab:robocasa}
\small
\begin{tabular}{@{}lc@{}}
\toprule
Method & Avg SR \\
\midrule
\multicolumn{2}{@{}l}{\textit{VLA-based methods}} \\
$\pi_0$~\cite{pi0} & 62.5 \\
GR00T-N1.5~\cite{gr00t} & 64.1 \\
\midrule
\multicolumn{2}{@{}l}{\textit{WAM-based methods}} \\
UWM~\cite{uwm} & 60.8 \\
DreamZero~\cite{dreamzero} & 62.4 \\
Cosmos Policy~\cite{cosmospolicy} & 67.1 \\
\midrule
\method (Ours) & \textbf{79.2} \\
\bottomrule
\end{tabular}
\end{minipage}
\hfill
\begin{minipage}[t]{0.48\textwidth}
\centering
\caption{Average success rate (\%) on 50 tasks of RoboTwin 2.0 benchmark.}
\label{tab:robotwin}
\small
\begin{tabular}{@{}lcc@{}}
\toprule
Method & Clean & Randomized \\
\midrule
\multicolumn{3}{@{}l}{\textit{VLA-based methods}} \\
$\pi_0$~\cite{pi0} & 65.9 & 58.4 \\
$\pi_{0.5}$~\cite{pi0.5} & 82.7 & 76.8 \\
\midrule
\multicolumn{3}{@{}l}{\textit{WAM-based methods}} \\
UWM~\cite{uwm} & 81.7 & 78.6 \\
GigaWorld-Policy~\cite{gigaworldpolicy} & 87.0 & 85.0 \\
Motus~\cite{motus} & 88.7 & 87.0 \\
\midrule
\method (Ours) & \textbf{89.8} & \textbf{90.7} \\
\bottomrule
\end{tabular}
\end{minipage}
\end{table}

\begin{table}[t]
\centering
\caption{4D reconstruction quality on RoboCasa. $\uparrow$ indicates higher is better; $\downarrow$ indicates lower is better.}
\label{tab:4d_recon}
\small
\begin{tabular}{@{}lcccccc@{}}
\toprule
\multirow{2}{*}{Method} & \multicolumn{3}{c}{RGB} & \multicolumn{2}{c}{Depth} & Point Cloud \\
\cmidrule(lr){2-4} \cmidrule(lr){5-6} \cmidrule(lr){7-7}
& PSNR$\uparrow$ & SSIM$\uparrow$ & LPIPS$\downarrow$ & AbsRel$\downarrow$ & $\delta_1$$\uparrow$ & CD$\downarrow$ \\
\midrule
DreamZero~\cite{dreamzero} + DA3~\cite{da3} & 21.12 & 0.7788 & 0.1580 & 0.1362 & 0.8594 & 0.0680 \\
Robot4DGen~\cite{robot4dgen} & 22.67 & 0.8207 & 0.1026 & 0.0736 & 0.9443 & 0.0134 \\
\method w/o depth + DA3~\cite{da3} & 23.09 & 0.8916 & 0.0548 & 0.1045 & 0.9089 & 0.0401 \\
\method (Ours) & \textbf{23.46} & \textbf{0.8942} & \textbf{0.0513} & \textbf{0.0349} & \textbf{0.9738} & \textbf{0.0049} \\
\bottomrule
\end{tabular}
\end{table}

As shown in Table~\ref{tab:4d_recon}, \method achieves the best performance across all metrics. Compared with the two-stage pipeline of DreamZero + DA3, \method improves PSNR by 2.34 dB and reduces Chamfer Distance from 0.0680 to 0.0049, demonstrating that end-to-end joint modeling produces substantially more accurate spatial reconstructions than post-hoc depth estimation applied to independently generated videos. Robot4DGen, which incorporates geometric priors during generation, achieves competitive depth metrics but falls short on visual fidelity (LPIPS 0.1026 vs. 0.0513). Notably, replacing our depth branch with Depth Anything 3 (\method w/o depth + DA3) preserves strong RGB quality but degrades depth accuracy (AbsRel 0.1045 vs. 0.0349) and point cloud quality (CD 0.0401 vs. 0.0049), confirming that the integrated depth branch produces more geometrically consistent predictions than a general-purpose monocular estimator.

\subsection{Ablation Studies}
\label{sec:exp_ablation}

We conduct ablation studies on the RoboCasa benchmark to validate the key design choices of \method. Due to computational constraints, all ablation variants are fine-tuned directly from the Wan2.2-TI2V-5B weights on the benchmark data without the large-scale pretraining stage described in Section~\ref{sec:training}. Unless otherwise stated, all variants share the same data, hyperparameters, and training schedule. Action latency is measured on a single NVIDIA RTX 3090 GPU.

\begin{table}[t]
\centering
\caption{Ablation studies on the RoboCasa benchmark. \textbf{Bold}: best; \underline{underline}: second best.}
\label{tab:ablation}
\footnotesize
\begin{tabular}{@{}lcccccccc@{}}
\toprule
\multirow{2}{*}{Variant} & \multirow{2}{*}{SR$\uparrow$} & Latency & \multicolumn{3}{c}{RGB} & \multicolumn{2}{c}{Depth} & Point Cloud \\
\cmidrule(lr){4-6} \cmidrule(lr){7-8} \cmidrule(lr){9-9}
& & (ms)$\downarrow$ & PSNR$\uparrow$ & SSIM$\uparrow$ & LPIPS$\downarrow$ & AbsRel$\downarrow$ & $\delta_1$$\uparrow$ & CD$\downarrow$ \\
\midrule
\multicolumn{9}{@{}l}{\textit{(a) Depth architecture design}} \\
No depth & 63.0 & \textbf{1033} & 23.09 & 0.8916 & 0.0548 & -- & -- & -- \\
Sequence concatenation & \textbf{68.7} & 1888 & \textbf{23.60} & \textbf{0.8987} & \textbf{0.0488} & \textbf{0.0332} & \textbf{0.9774} & \textbf{0.0037} \\
Channel concatenation & 64.2 & 1266 & 23.20 & 0.8933 & 0.0522 & 0.0377 & 0.9728 & 0.0052 \\
Interleaved branch (Ours) & \underline{67.8} & \textbf{1033} & \underline{23.46} & \underline{0.8942} & \underline{0.0513} & \underline{0.0349} & \underline{0.9738} & \underline{0.0049} \\
\midrule
\multicolumn{9}{@{}l}{\textit{(b) Noise scheduling strategy}} \\
Sync train + Sync infer & 66.4 & 4665 & \textbf{23.48} & \textbf{0.8963} & \textbf{0.0502} & \underline{0.0375} & \underline{0.9725} & \textbf{0.0041} \\
Decoupled train + Sync infer & 66.3 & 4665 & 23.17 & 0.8937 & 0.0529 & 0.0397 & 0.9701 & 0.0050 \\
Decoupled train + Async infer & \underline{67.2} & \textbf{1033} & 22.60 & 0.8878 & 0.0561 & 0.0430 & 0.9680 & 0.0061 \\
ANS train + Async infer (Ours) & \textbf{67.8} & \textbf{1033} & \underline{23.46} & \underline{0.8942} & \underline{0.0513} & \textbf{0.0349} & \textbf{0.9738} & \underline{0.0049} \\
\bottomrule
\end{tabular}
\end{table}

\paragraph{Depth architecture design.} We compare four depth incorporation strategies (Table~\ref{tab:ablation}(a)). The Latency column reports the action generation latency during policy execution. Sequence concatenation achieves the best quality metrics by treating depth as explicit tokens, but nearly doubles the action latency to 1888\,ms due to the expanded sequence length. Channel concatenation also introduces noticeable overhead (1266\,ms). In contrast, our interleaved branch matches the latency of the no-depth variant (1033\,ms), since the depth branch can be toggled off during action decoding, while delivering clearly superior quality over both the no-depth and channel-concatenation variants. Notably, removing depth supervision entirely causes the policy success rate to drop from 67.8\% to 63.0\%, confirming that explicit spatial modeling is essential for robust manipulation. Channel concatenation also underperforms in success rate (64.2\%), as fusing depth along the channel dimension shifts the input distribution away from the pretrained manifold.

\paragraph{Effect of ANS.} We compare four noise scheduling configurations (Table~\ref{tab:ablation}(b)), where synchronous variants use 25 joint steps and asynchronous variants decode actions in only 5 steps. The synchronous baselines yield strong RGB metrics but force the policy to wait for all 25 denoising steps, resulting in an action latency of 4665\,ms. Asynchronous inference reduces this to 1033\,ms, a $4.5\times$ speedup, by decoding actions in only the first 5 steps. Among the asynchronous variants, Decoupled-Async achieves a competitive success rate (67.2\%) but its reconstruction quality degrades significantly (PSNR 22.60, AbsRel 0.0430), because the video branch must continue denoising conditioned on clean actions, a regime never seen during independently sampled training. Our ANS closes this gap by coupling the training noise distribution to faithfully cover the asynchronous inference regime. As a result, ANS achieves the highest success rate (67.8\%) and the best depth metrics at the same 1033\,ms latency, while maintaining RGB quality competitive with the synchronous baseline.

\section{Conclusion}

In this work, we presented \method, a unified 4D World Action Model that extends unified world action modeling into spatially aware 4D dynamics simulation. Through a lightweight depth adaptation module that replicates the final DiT blocks as an interleaved depth branch, \method achieves high-quality spatial reconstruction without increasing sequence length or compromising pretrained visual priors. Asynchronous Noise Sampling further aligns training and inference noise distributions across modalities, enabling rapid action decoding while preserving video generation quality. Experiments on RoboCasa and RoboTwin 2.0 demonstrate that \method consistently outperforms all baselines in both policy success rate and 4D reconstruction quality, confirming that a single framework can jointly optimize policy execution, visual generation, and spatial reconstruction.

\bibliographystyle{unsrtnat}
\bibliography{citations}

@article{3dgs,
  author       = {Bernhard Kerbl and
                  Georgios Kopanas and
                  Thomas Leimk{\"{u}}hler and
                  George Drettakis},
  title        = {3D Gaussian Splatting for Real-Time Radiance Field Rendering},
  journal      = {{ACM} Trans. Graph.},
  volume       = {42},
  number       = {4},
  pages        = {139:1--139:14},
  year         = {2023},
}

@inproceedings{3dvla,
  author       = {Haoyu Zhen and
                  Xiaowen Qiu and
                  Peihao Chen and
                  Jincheng Yang and
                  Xin Yan and
                  Yilun Du and
                  Yining Hong and
                  Chuang Gan},
  title        = {3D-VLA: {A} 3D Vision-Language-Action Generative World Model},
  booktitle    = {Forty-first International Conference on Machine Learning},
  pages        = {61229--61245},
  year         = {2024},
}

@article{bagelvla,
  author       = {Yucheng Hu and
                  Jianke Zhang and
                  Yuanfei Luo and
                  Yanjiang Guo and
                  Xiaoyu Chen and
                  Xinshu Sun and
                  Kun Feng and
                  Qingzhou Lu and
                  Sheng Chen and
                  Yangang Zhang and
                  Wei Li and
                  Jianyu Chen},
  title        = {BagelVLA: Enhancing Long-Horizon Manipulation via Interleaved Vision-Language-Action
                  Generation},
  journal      = {CoRR},
  volume       = {abs/2602.09849},
  year         = {2026},
}

@article{bridgevla,
  author       = {Peiyan Li and
                  Yixiang Chen and
                  Hongtao Wu and
                  Xiao Ma and
                  Xiangnan Wu and
                  Yan Huang and
                  Liang Wang and
                  Tao Kong and
                  Tieniu Tan},
  title        = {BridgeVLA: Input-Output Alignment for Efficient 3D Manipulation Learning
                  with Vision-Language Models},
  journal      = {CoRR},
  volume       = {abs/2506.07961},
  year         = {2025},
}

@article{cosmospolicy,
  author       = {Moo Jin Kim and
                  Yihuai Gao and
                  Tsung{-}Yi Lin and
                  Yen{-}Chen Lin and
                  Yunhao Ge and
                  Grace Lam and
                  Percy Liang and
                  Shuran Song and
                  Ming{-}Yu Liu and
                  Chelsea Finn and
                  Jinwei Gu},
  title        = {Cosmos Policy: Fine-Tuning Video Models for Visuomotor Control and
                  Planning},
  journal      = {CoRR},
  volume       = {abs/2601.16163},
  year         = {2026},
}

@article{da3,
  author       = {Haotong Lin and
                  Sili Chen and
                  Junhao Liew and
                  Donny Y. Chen and
                  Zhenyu Li and
                  Guang Shi and
                  Jiashi Feng and
                  Bingyi Kang},
  title        = {Depth Anything 3: Recovering the Visual Space from Any Views},
  journal      = {CoRR},
  volume       = {abs/2511.10647},
  year         = {2025},
}

@article{dit4dit,
  author       = {Teli Ma and
                  Jia Zheng and
                  Zifan Wang and
                  Chunli Jiang and
                  Andy Cui and
                  Junwei Liang and
                  Shuo Yang},
  title        = {DiT4DiT: Jointly Modeling Video Dynamics and Actions for Generalizable
                  Robot Control},
  journal      = {CoRR},
  volume       = {abs/2603.10448},
  year         = {2026},
}

@article{dreamvla,
  author       = {Wenyao Zhang and
                  Hongsi Liu and
                  Zekun Qi and
                  Yunnan Wang and
                  Xinqiang Yu and
                  Jiazhao Zhang and
                  Runpei Dong and
                  Jiawei He and
                  He Wang and
                  Zhizheng Zhang and
                  Li Yi and
                  Wenjun Zeng and
                  Xin Jin},
  title        = {DreamVLA: {A} Vision-Language-Action Model Dreamed with Comprehensive
                  World Knowledge},
  journal      = {CoRR},
  volume       = {abs/2507.04447},
  year         = {2025},
}

@article{dreamzero,
  author       = {Seonghyeon Ye and
                  Yunhao Ge and
                  Kaiyuan Zheng and
                  Shenyuan Gao and
                  Sihyun Yu and
                  George Kurian and
                  Suneel Indupuru and
                  You Liang Tan and
                  Chuning Zhu and
                  Jiannan Xiang and
                  Ayaan Malik and
                  Kyungmin Lee and
                  William Liang and
                  Nadun Ranawaka and
                  Jiasheng Gu and
                  Yinzhen Xu and
                  Guanzhi Wang and
                  Fengyuan Hu and
                  Avnish Narayan and
                  Johan Bjorck and
                  Jing Wang and
                  Gwanghyun Kim and
                  Dantong Niu and
                  Ruijie Zheng and
                  Yuqi Xie and
                  Jimmy Wu and
                  Qi Wang and
                  Ryan Julian and
                  Danfei Xu and
                  Yilun Du and
                  Yevgen Chebotar and
                  Scott Reed and
                  Jan Kautz and
                  Yuke Zhu and
                  Linxi "Jim" Fan and
                  Joel Jang},
  title        = {World Action Models are Zero-shot Policies},
  journal      = {CoRR},
  volume       = {abs/2602.15922},
  year         = {2026},
}

@inproceedings{dust3r,
  author       = {Shuzhe Wang and
                  Vincent Leroy and
                  Yohann Cabon and
                  Boris Chidlovskii and
                  J{\'{e}}r{\^{o}}me Revaud},
  title        = {DUSt3R: Geometric 3D Vision Made Easy},
  booktitle    = {{IEEE/CVF} Conference on Computer Vision and Pattern Recognition},
  pages        = {20697--20709},
  year         = {2024},
}

@article{enerverse,
  author       = {Siyuan Huang and
                  Liliang Chen and
                  Pengfei Zhou and
                  Shengcong Chen and
                  Zhengkai Jiang and
                  Yutao Hu and
                  Peng Gao and
                  Hongsheng Li and
                  Maoqing Yao and
                  Guanghui Ren},
  title        = {EnerVerse: Envisioning Embodied Future Space for Robotics Manipulation},
  journal      = {CoRR},
  volume       = {abs/2501.01895},
  year         = {2025},
}

@article{evo0,
  author       = {Tao Lin and
                  Gen Li and
                  Yilei Zhong and
                  Yanwen Zou and
                  Bo Zhao},
  title        = {Evo-0: Vision-Language-Action Model with Implicit Spatial Understanding},
  journal      = {CoRR},
  volume       = {abs/2507.00416},
  year         = {2025},
}

@article{falcon,
  author       = {Zhengshen Zhang and
                  Hao Li and
                  Yalun Dai and
                  Zhengbang Zhu and
                  Lei Zhou and
                  Chenchen Liu and
                  Dong Wang and
                  Francis E. H. Tay and
                  Sijin Chen and
                  Ziwei Liu and
                  Yuxiao Liu and
                  Xinghang Li and
                  Pan Zhou},
  title        = {From Spatial to Actions: Grounding Vision-Language-Action Model in
                  Spatial Foundation Priors},
  journal      = {CoRR},
  volume       = {abs/2510.17439},
  year         = {2025},
}

@article{fastwam,
  author       = {Tianyuan Yuan and
                  Zibin Dong and
                  Yicheng Liu and
                  Hang Zhao},
  title        = {Fast-WAM: Do World Action Models Need Test-time Future Imagination?},
  journal      = {CoRR},
  volume       = {abs/2603.16666},
  year         = {2026},
}

@article{flowdreamer,
  author       = {Jun Guo and
                  Xiaojian Ma and
                  Yikai Wang and
                  Min Yang and
                  Huaping Liu and
                  Qing Li},
  title        = {FlowDreamer: {A} {RGB-D} World Model With Flow-Based Motion Representations
                  for Robot Manipulation},
  journal      = {{IEEE} Robotics Automation Letters},
  volume       = {11},
  number       = {3},
  pages        = {2466--2473},
  year         = {2026},
}

@article{genieenvisioner,
  author       = {Yue Liao and
                  Pengfei Zhou and
                  Siyuan Huang and
                  Donglin Yang and
                  Shengcong Chen and
                  Yuxin Jiang and
                  Hu Yue and
                  Jingbin Cai and
                  Si Liu and
                  Jianlan Luo and
                  Liliang Chen and
                  Shuicheng Yan and
                  Maoqing Yao and
                  Guanghui Ren},
  title        = {Genie Envisioner: {A} Unified World Foundation Platform for Robotic
                  Manipulation},
  journal      = {CoRR},
  volume       = {abs/2508.05635},
  year         = {2025},
}

@article{geovla,
  author       = {Lin Sun and
                  Bin Xie and
                  Yingfei Liu and
                  Hao Shi and
                  Tiancai Wang and
                  Jiale Cao},
  title        = {GeoVLA: Empowering 3D Representations in Vision-Language-Action Models},
  journal      = {CoRR},
  volume       = {abs/2508.09071},
  year         = {2025},
}

@article{gigaworldpolicy,
  author       = {Angen Ye and
                  Boyuan Wang and
                  Chaojun Ni and
                  Guan Huang and
                  Guosheng Zhao and
                  Hao Li and
                  Hengtao Li and
                  Jie Li and
                  Jindi Lv and
                  Jingyu Liu and
                  Min Cao and
                  Peng Li and
                  Qiuping Deng and
                  Wenjun Mei and
                  Xiaofeng Wang and
                  Xinze Chen and
                  Xinyu Zhou and
                  Yang Wang and
                  Yifan Chang and
                  Yifan Li and
                  Yukun Zhou and
                  Yun Ye and
                  Zhichao Liu and
                  Zheng Zhu},
  title        = {GigaWorld-Policy: An Efficient Action-Centered World-Action Model},
  journal      = {CoRR},
  volume       = {abs/2603.17240},
  year         = {2026},
}

@article{gr00t,
  author       = {Johan Bjorck and
                  Fernando Casta{\~{n}}eda and
                  Nikita Cherniadev and
                  Xingye Da and
                  Runyu Ding and
                  Linxi Fan and
                  Yu Fang and
                  Dieter Fox and
                  Fengyuan Hu and
                  Spencer Huang and
                  Joel Jang and
                  Zhenyu Jiang and
                  Jan Kautz and
                  Kaushil Kundalia and
                  Lawrence Lao and
                  Zhiqi Li and
                  Zongyu Lin and
                  Kevin Lin and
                  Guilin Liu and
                  Edith LLontop and
                  Loic Magne and
                  Ajay Mandlekar and
                  Avnish Narayan and
                  Soroush Nasiriany and
                  Scott Reed and
                  You Liang Tan and
                  Guanzhi Wang and
                  Zu Wang and
                  Jing Wang and
                  Qi Wang and
                  Jiannan Xiang and
                  Yuqi Xie and
                  Yinzhen Xu and
                  Zhenjia Xu and
                  Seonghyeon Ye and
                  Zhiding Yu and
                  Ao Zhang and
                  Hao Zhang and
                  Yizhou Zhao and
                  Ruijie Zheng and
                  Yuke Zhu},
  title        = {{GR00T} {N1:} An Open Foundation Model for Generalist Humanoid Robots},
  journal      = {CoRR},
  volume       = {abs/2503.14734},
  year         = {2025},
}

@inproceedings{gr1,
  author       = {Hongtao Wu and
                  Ya Jing and
                  Chilam Cheang and
                  Guangzeng Chen and
                  Jiafeng Xu and
                  Xinghang Li and
                  Minghuan Liu and
                  Hang Li and
                  Tao Kong},
  title        = {Unleashing Large-Scale Video Generative Pre-training for Visual Robot
                  Manipulation},
  booktitle    = {The Twelfth International Conference on Learning Representations},
  year         = {2024},
}

@article{gwm,
  author       = {Guanxing Lu and
                  Baoxiong Jia and
                  Puhao Li and
                  Yixin Chen and
                  Ziwei Wang and
                  Yansong Tang and
                  Siyuan Huang},
  title        = {{GWM:} Towards Scalable Gaussian World Models for Robotic Manipulation},
  journal      = {CoRR},
  volume       = {abs/2508.17600},
  year         = {2025},
}

@article{lingbotva,
  author       = {Lin Li and
                  Qihang Zhang and
                  Yiming Luo and
                  Shuai Yang and
                  Ruilin Wang and
                  Fei Han and
                  Mingrui Yu and
                  Zelin Gao and
                  Nan Xue and
                  Xing Zhu and
                  Yujun Shen and
                  Yinghao Xu},
  title        = {Causal World Modeling for Robot Control},
  journal      = {CoRR},
  volume       = {abs/2601.21998},
  year         = {2026},
}

@inproceedings{manigaussian,
  author       = {Guanxing Lu and
                  Shiyi Zhang and
                  Ziwei Wang and
                  Changliu Liu and
                  Jiwen Lu and
                  Yansong Tang},
  title        = {ManiGaussian: Dynamic Gaussian Splatting for Multi-task Robotic Manipulation},
  booktitle    = {18th European Conference on Computer Vision},
  pages        = {349--366},
  year         = {2024},
}

@article{mimicvideo,
  author       = {Jonas Pai and
                  Liam Achenbach and
                  Victoriano Montesinos and
                  Benedek Forrai and
                  Oier Mees and
                  Elvis Nava},
  title        = {mimic-video: Video-Action Models for Generalizable Robot Control Beyond
                  VLAs},
  journal      = {CoRR},
  volume       = {abs/2512.15692},
  year         = {2025},
}

@article{moto,
  author       = {Yi Chen and
                  Yuying Ge and
                  Yizhuo Li and
                  Yixiao Ge and
                  Mingyu Ding and
                  Ying Shan and
                  Xihui Liu},
  title        = {Moto: Latent Motion Token as the Bridging Language for Robot Manipulation},
  journal      = {CoRR},
  volume       = {abs/2412.04445},
  year         = {2024},
}

@article{motus,
  author       = {Hongzhe Bi and
                  Hengkai Tan and
                  Shenghao Xie and
                  Zeyuan Wang and
                  Shuhe Huang and
                  Haitian Liu and
                  Ruowen Zhao and
                  Yao Feng and
                  Chendong Xiang and
                  Yinze Rong and
                  Hongyan Zhao and
                  Hanyu Liu and
                  Zhizhong Su and
                  Lei Ma and
                  Hang Su and
                  Jun Zhu},
  title        = {Motus: {A} Unified Latent Action World Model},
  journal      = {CoRR},
  volume       = {abs/2512.13030},
  year         = {2025},
}

@article{mvista4d,
  author       = {Jiaxu Wang and
                  Yicheng Jiang and
                  Tianlun He and
                  Jingkai Sun and
                  Qiang Zhang and
                  Junhao He and
                  Jiahang Cao and
                  Zesen Gan and
                  Mingyuan Sun and
                  Qiming Shao and
                  Xiangyu Yue},
  title        = {{MVISTA-4D:} View-Consistent 4D World Model with Test-Time Action
                  Inference for Robotic Manipulation},
  journal      = {CoRR},
  volume       = {abs/2602.09878},
  year         = {2026},
}

@inproceedings{octo,
  author       = {Dibya Ghosh and
                  Homer Rich Walke and
                  Karl Pertsch and
                  Kevin Black and
                  Oier Mees and
                  Sudeep Dasari and
                  Joey Hejna and
                  Tobias Kreiman and
                  Charles Xu and
                  Jianlan Luo and
                  You Liang Tan and
                  Lawrence Yunliang Chen and
                  Quan Vuong and
                  Ted Xiao and
                  Pannag R. Sanketi and
                  Dorsa Sadigh and
                  Chelsea Finn and
                  Sergey Levine},
  title        = {Octo: An Open-Source Generalist Robot Policy},
  booktitle    = {Robotics: Science and Systems},
  year         = {2024},
}

@inproceedings{openvla,
  author       = {Moo Jin Kim and
                  Karl Pertsch and
                  Siddharth Karamcheti and
                  Ted Xiao and
                  Ashwin Balakrishna and
                  Suraj Nair and
                  Rafael Rafailov and
                  Ethan Paul Foster and
                  Pannag R. Sanketi and
                  Quan Vuong and
                  Thomas Kollar and
                  Benjamin Burchfiel and
                  Russ Tedrake and
                  Dorsa Sadigh and
                  Sergey Levine and
                  Percy Liang and
                  Chelsea Finn},
  title        = {OpenVLA: An Open-Source Vision-Language-Action Model},
  booktitle    = {Conference on Robot Learning},
  pages        = {2679--2713},
  year         = {2024},
}

@article{pi0,
  author       = {Kevin Black and
                  Noah Brown and
                  Danny Driess and
                  Adnan Esmail and
                  Michael Equi and
                  Chelsea Finn and
                  Niccolo Fusai and
                  Lachy Groom and
                  Karol Hausman and
                  Brian Ichter and
                  Szymon Jakubczak and
                  Tim Jones and
                  Liyiming Ke and
                  Sergey Levine and
                  Adrian Li{-}Bell and
                  Mohith Mothukuri and
                  Suraj Nair and
                  Karl Pertsch and
                  Lucy Xiaoyang Shi and
                  James Tanner and
                  Quan Vuong and
                  Anna Walling and
                  Haohuan Wang and
                  Ury Zhilinsky},
  title        = {{\(\pi\)}\({}_{\mbox{0}}\): {A} Vision-Language-Action Flow Model
                  for General Robot Control},
  journal      = {CoRR},
  volume       = {abs/2410.24164},
  year         = {2024},
}

@article{pi0.5,
  author       = {Physical Intelligence and
                  Kevin Black and
                  Noah Brown and
                  James Darpinian and
                  Karan Dhabalia and
                  Danny Driess and
                  Adnan Esmail and
                  Michael Equi and
                  Chelsea Finn and
                  Niccolo Fusai and
                  Manuel Y. Galliker and
                  Dibya Ghosh and
                  Lachy Groom and
                  Karol Hausman and
                  Brian Ichter and
                  Szymon Jakubczak and
                  Tim Jones and
                  Liyiming Ke and
                  Devin LeBlanc and
                  Sergey Levine and
                  Adrian Li{-}Bell and
                  Mohith Mothukuri and
                  Suraj Nair and
                  Karl Pertsch and
                  Allen Z. Ren and
                  Lucy Xiaoyang Shi and
                  Laura Smith and
                  Jost Tobias Springenberg and
                  Kyle Stachowicz and
                  James Tanner and
                  Quan Vuong and
                  Homer Walke and
                  Anna Walling and
                  Haohuan Wang and
                  Lili Yu and
                  Ury Zhilinsky},
  title        = {{\(\pi\)}\({}_{\mbox{0.5}}\): a Vision-Language-Action Model with
                  Open-World Generalization},
  journal      = {CoRR},
  volume       = {abs/2504.16054},
  year         = {2025},
}

@article{pi0.7,
  title={$\pi_{0.7}$: a Steerable Generalist Robotic Foundation Model with Emergent Capabilities},
  author={Intelligence, Physical and Ai, Bo and Amin, Ali and Aniceto, Raichelle and Balakrishna, Ashwin and Balke, Greg and Black, Kevin and Bokinsky, George and Cao, Shihao and Charbonnier, Thomas and others},
  journal={arXiv preprint arXiv:2604.15483},
  year={2026}
}

@article{pointvla,
  author       = {Chengmeng Li and
                  Junjie Wen and
                  Yaxin Peng and
                  Yan Peng and
                  Yichen Zhu},
  title        = {PointVLA: Injecting the 3D World Into Vision-Language-Action Models},
  journal      = {{IEEE} Robotics Autom. Lett.},
  volume       = {11},
  number       = {3},
  pages        = {2506--2513},
  year         = {2026},
}

@article{pointworld,
  author       = {Wenbing Huang and
                  Yu{-}Wei Chao and
                  Arsalan Mousavian and
                  Ming{-}Yu Liu and
                  Dieter Fox and
                  Kaichun Mo and
                  Li Fei{-}Fei},
  title        = {PointWorld: Scaling 3D World Models for In-The-Wild Robotic Manipulation},
  journal      = {CoRR},
  volume       = {abs/2601.03782},
  year         = {2026},
}

@inproceedings{rdt,
  author       = {Songming Liu and
                  Lingxuan Wu and
                  Bangguo Li and
                  Hengkai Tan and
                  Huayu Chen and
                  Zhengyi Wang and
                  Ke Xu and
                  Hang Su and
                  Jun Zhu},
  title        = {{RDT-1B:} a Diffusion Foundation Model for Bimanual Manipulation},
  booktitle    = {The Thirteenth International Conference on Learning Representations},
  year         = {2025},
}

@inproceedings{roboflamingo,
  author       = {Xinghang Li and
                  Minghuan Liu and
                  Hanbo Zhang and
                  Cunjun Yu and
                  Jie Xu and
                  Hongtao Wu and
                  Chilam Cheang and
                  Ya Jing and
                  Weinan Zhang and
                  Huaping Liu and
                  Hang Li and
                  Tao Kong},
  title        = {Vision-Language Foundation Models as Effective Robot Imitators},
  booktitle    = {The Twelfth International Conference on Learning Representations},
  year         = {2024},
}

@article{robot4dgen,
  author       = {Zeyi Liu and
                  Shuang Li and
                  Eric Cousineau and
                  Siyuan Feng and
                  Benjamin Burchfiel and
                  Shuran Song},
  title        = {Geometry-aware 4D Video Generation for Robot Manipulation},
  journal      = {CoRR},
  volume       = {abs/2507.01099},
  year         = {2025},
}

@inproceedings{rt2,
  author       = {Brianna Zitkovich and
                  Tianhe Yu and
                  Sichun Xu and
                  Peng Xu and
                  Ted Xiao and
                  Fei Xia and
                  Jialin Wu and
                  Paul Wohlhart and
                  Stefan Welker and
                  Ayzaan Wahid and
                  Quan Vuong and
                  Vincent Vanhoucke and
                  Huong T. Tran and
                  Radu Soricut and
                  Anikait Singh and
                  Jaspiar Singh and
                  Pierre Sermanet and
                  Pannag R. Sanketi and
                  Grecia Salazar and
                  Michael S. Ryoo and
                  Krista Reymann and
                  Kanishka Rao and
                  Karl Pertsch and
                  Igor Mordatch and
                  Henryk Michalewski and
                  Yao Lu and
                  Sergey Levine and
                  Lisa Lee and
                  Tsang{-}Wei Edward Lee and
                  Isabel Leal and
                  Yuheng Kuang and
                  Dmitry Kalashnikov and
                  Ryan Julian and
                  Nikhil J. Joshi and
                  Alex Irpan and
                  Brian Ichter and
                  Jasmine Hsu and
                  Alexander Herzog and
                  Karol Hausman and
                  Keerthana Gopalakrishnan and
                  Chuyuan Fu and
                  Pete Florence and
                  Chelsea Finn and
                  Kumar Avinava Dubey and
                  Danny Driess and
                  Tianli Ding and
                  Krzysztof Marcin Choromanski and
                  Xi Chen and
                  Yevgen Chebotar and
                  Justice Carbajal and
                  Noah Brown and
                  Anthony Brohan and
                  Montserrat Gonzalez Arenas and
                  Kehang Han},
  title        = {{RT-2:} Vision-Language-Action Models Transfer Web Knowledge to Robotic
                  Control},
  booktitle    = {Conference on Robot Learning},
  pages        = {2165--2183},
  year         = {2023},
}

@article{spatialforcing,
  author       = {Fuhao Li and
                  Wenxuan Song and
                  Han Zhao and
                  Jingbo Wang and
                  Pengxiang Ding and
                  Donglin Wang and
                  Long Zeng and
                  Haoang Li},
  title        = {Spatial Forcing: Implicit Spatial Representation Alignment for Vision-language-action
                  Model},
  journal      = {CoRR},
  volume       = {abs/2510.12276},
  year         = {2025},
}

@article{spatial-vla,
  author       = {Delin Qu and
                  Haoming Song and
                  Qizhi Chen and
                  Yuanqi Yao and
                  Xinyi Ye and
                  Yan Ding and
                  Zhigang Wang and
                  JiaYuan Gu and
                  Bin Zhao and
                  Dong Wang and
                  Xuelong Li},
  title        = {SpatialVLA: Exploring Spatial Representations for Visual-Language-Action
                  Model},
  journal      = {CoRR},
  volume       = {abs/2501.15830},
  year         = {2025},
}

@article{tesseract,
  author       = {Haoyu Zhen and
                  Qiao Sun and
                  Hongxin Zhang and
                  Junyan Li and
                  Siyuan Zhou and
                  Yilun Du and
                  Chuang Gan},
  title        = {TesserAct: Learning 4D Embodied World Models},
  journal      = {CoRR},
  volume       = {abs/2504.20995},
  year         = {2025},
}

@inproceedings{unipi,
  author       = {Yilun Du and
                  Sherry Yang and
                  Bo Dai and
                  Hanjun Dai and
                  Ofir Nachum and
                  Josh Tenenbaum and
                  Dale Schuurmans and
                  Pieter Abbeel},
  title        = {Learning Universal Policies via Text-Guided Video Generation},
  booktitle    = {Advances in Neural Information Processing Systems},
  year         = {2023},
}

@article{univlahang,
  author       = {Yuqi Wang and
                  Xinghang Li and
                  Wenxuan Wang and
                  Junbo Zhang and
                  Yingyan Li and
                  Yuntao Chen and
                  Xinlong Wang and
                  Zhaoxiang Zhang},
  title        = {Unified Vision-Language-Action Model},
  journal      = {CoRR},
  volume       = {abs/2506.19850},
  year         = {2025},
}

@inproceedings{vda,
  author       = {Sili Chen and
                  Hengkai Guo and
                  Shengnan Zhu and
                  Feihu Zhang and
                  Zilong Huang and
                  Jiashi Feng and
                  Bingyi Kang},
  title        = {Video Depth Anything: Consistent Depth Estimation for Super-Long Videos},
  booktitle    = {{IEEE/CVF} Conference on Computer Vision and Pattern Recognition},
  pages        = {22831--22840},
  year         = {2025},
}

@inproceedings{vggt,
  author       = {Jianyuan Wang and
                  Minghao Chen and
                  Nikita Karaev and
                  Andrea Vedaldi and
                  Christian Rupprecht and
                  David Novotn{\'{y}}},
  title        = {{VGGT:} Visual Geometry Grounded Transformer},
  booktitle    = {{IEEE/CVF} Conference on Computer Vision and Pattern Recognition},
  pages        = {5294--5306},
  year         = {2025},
}

@article{videovla,
  author       = {Yichao Shen and
                  Fangyun Wei and
                  Zhiying Du and
                  Yaobo Liang and
                  Yan Lu and
                  Jiaolong Yang and
                  Nanning Zheng and
                  Baining Guo},
  title        = {VideoVLA: Video Generators Can Be Generalizable Robot Manipulators},
  journal      = {CoRR},
  volume       = {abs/2512.06963},
  year         = {2025},
}

@article{vista,
  author       = {Qian Long and
                  Yueze Wang and
                  Jiaxi Song and
                  Junbo Zhang and
                  Peiyan Li and
                  Wenxuan Wang and
                  Yuqi Wang and
                  Haoyang Li and
                  Shaoxuan Xie and
                  Guocai Yao and
                  Hanbo Zhang and
                  Xinlong Wang and
                  Zhongyuan Wang and
                  Xuguang Lan and
                  Huaping Liu and
                  Xinghang Li},
  title        = {Scaling World Model for Hierarchical Manipulation Policies},
  journal      = {CoRR},
  volume       = {abs/2602.10983},
  year         = {2026},
}

@article{vlajepa,
  author       = {Jingwen Sun and
                  Wenyao Zhang and
                  Zekun Qi and
                  Shaojie Ren and
                  Zezhi Liu and
                  Hanxin Zhu and
                  Guangzhong Sun and
                  Xin Jin and
                  Zhibo Chen},
  title        = {{VLA-JEPA:} Enhancing Vision-Language-Action Model with Latent World
                  Model},
  journal      = {CoRR},
  volume       = {abs/2602.10098},
  year         = {2026},
}

@inproceedings{vpp,
  author       = {Yucheng Hu and
                  Yanjiang Guo and
                  Pengchao Wang and
                  Xiaoyu Chen and
                  Yen{-}Jen Wang and
                  Jianke Zhang and
                  Koushil Sreenath and
                  Chaochao Lu and
                  Jianyu Chen},
  title        = {Video Prediction Policy: {A} Generalist Robot Policy with Predictive
                  Visual Representations},
  booktitle    = {Forty-second International Conference on Machine Learning},
  year         = {2025},
}

@article{wamsurvey,
  author       = {Zhanguang Zhang and
                  Zhiyuan Li and
                  Behnam Rahmati and
                  Rui Heng Yang and
                  Yintao Ma and
                  Amir Rasouli and
                  Sajjad Pakdamansavoji and
                  Yangzheng Wu and
                  Lingfeng Zhang and
                  Tongtong Cao and
                  Feng Wen and
                  Xinyu Wang and
                  Xingyue Quan and
                  Yingxue Zhang},
  title        = {Do World Action Models Generalize Better than VLAs? {A} Robustness
                  Study},
  journal      = {CoRR},
  volume       = {abs/2603.22078},
  year         = {2026},
}

@article{worldvla,
  author       = {Jun Cen and
                  Chaohui Yu and
                  Hangjie Yuan and
                  Yuming Jiang and
                  Siteng Huang and
                  Jiayan Guo and
                  Xin Li and
                  Yibing Song and
                  Hao Luo and
                  Fan Wang and
                  Deli Zhao and
                  Hao Chen},
  title        = {WorldVLA: Towards Autoregressive Action World Model},
  journal      = {CoRR},
  volume       = {abs/2506.21539},
  year         = {2025},
}

@article{wristworld,
  author       = {Zezhong Qian and
                  Xiaowei Chi and
                  Yuming Li and
                  Shizun Wang and
                  Zhiyuan Qin and
                  Xiaozhu Ju and
                  Sirui Han and
                  Shanghang Zhang},
  title        = {WristWorld: Generating Wrist-Views via 4D World Models for Robotic
                  Manipulation},
  journal      = {CoRR},
  volume       = {abs/2510.07313},
  year         = {2025},
}

@article{xr0,
  author       = {Rui Cai and
                  Jun Guo and
                  Xinze He and
                  Piaopiao Jin and
                  Jie Li and
                  Bingxuan Lin and
                  Futeng Liu and
                  Wei Liu and
                  Fei Ma and
                  Kun Ma and
                  Feng Qiu and
                  Heng Qu and
                  Yifei Su and
                  Qiao Sun and
                  Dong Wang and
                  Donghao Wang and
                  Yunhong Wang and
                  Rujie Wu and
                  Diyun Xiang and
                  Yu Yang and
                  Hangjun Ye and
                  Yuan Zhang and
                  Quanyun Zhou},
  title        = {Xiaomi-Robotics-0: An Open-Sourced Vision-Language-Action Model with
                  Real-Time Execution},
  journal      = {CoRR},
  volume       = {abs/2602.12684},
  year         = {2026},
}

@article{uwm,
  author       = {Chuning Zhu and
                  Raymond Yu and
                  Siyuan Feng and
                  Benjamin Burchfiel and
                  Paarth Shah and
                  Abhishek Gupta},
  title        = {Unified World Models: Coupling Video and Action Diffusion for Pretraining
                  on Large Robotic Datasets},
  journal      = {CoRR},
  volume       = {abs/2504.02792},
  year         = {2025},
}

@article{diffusionpolicy,
  title={Diffusion policy: Visuomotor policy learning via action diffusion},
  author={Chi, Cheng and Xu, Zhenjia and Feng, Siyuan and Cousineau, Eric and Du, Yilun and Burchfiel, Benjamin and Tedrake, Russ and Song, Shuran},
  journal={The International Journal of Robotics Research},
  volume={44},
  number={10-11},
  pages={1684--1704},
  year={2025},
}

@inproceedings{ddim,
  title={Denoising Diffusion Implicit Models},
  author={Song, Jiaming and Meng, Chenlin and Ermon, Stefano},
  booktitle={International Conference on Learning Representations},
  year={2021}
}

@inproceedings{edm,
  title={Elucidating the design space of diffusion-based generative models},
  author={Karras, Tero and Aittala, Miika and Aila, Timo and Laine, Samuli},
  booktitle={Advances in neural information processing systems},
  volume={35},
  pages={26565--26577},
  year={2022}
}

@inproceedings{unipc,
  title={Unipc: A unified predictor-corrector framework for fast sampling of diffusion models},
  author={Zhao, Wenliang and Bai, Lujia and Rao, Yongming and Zhou, Jie and Lu, Jiwen},
  booktitle={Advances in Neural Information Processing Systems},
  volume={36},
  pages={49842--49869},
  year={2023}
}

@article{wan,
  title={Wan: Open and advanced large-scale video generative models},
  author={Wan, Team and Wang, Ang and Ai, Baole and Wen, Bin and Mao, Chaojie and Xie, Chen-Wei and Chen, Di and Yu, Feiwu and Zhao, Haiming and Yang, Jianxiao and others},
  journal={arXiv preprint arXiv:2503.20314},
  year={2025}
}

@inproceedings{robocasa,
  title={RoboCasa: Large-Scale Simulation of Everyday Tasks for Generalist Robots},
  author={Nasiriany, Soroush and Maddukuri, Abhiram and Zhang, Lance and Parikh, Adeet and Lo, Aaron and Joshi, Abhishek and Mandlekar, Ajay and Zhu, Yuke},
  booktitle={RSS Workshop: Data Generation for Robotics},
  year={2024}
}

@article{robotwin2,
  title={Robotwin 2.0: A scalable data generator and benchmark with strong domain randomization for robust bimanual robotic manipulation},
  author={Chen, Tianxing and Chen, Zanxin and Chen, Baijun and Cai, Zijian and Liu, Yibin and Li, Zixuan and Liang, Qiwei and Lin, Xianliang and Ge, Yiheng and Gu, Zhenyu and others},
  journal={arXiv preprint arXiv:2506.18088},
  year={2025}
}

@article{rope,
  title={Roformer: Enhanced transformer with rotary position embedding},
  author={Su, Jianlin and Ahmed, Murtadha and Lu, Yu and Pan, Shengfeng and Bo, Wen and Liu, Yunfeng},
  journal={Neurocomputing},
  volume={568},
  pages={127063},
  year={2024},
}

@inproceedings{flowmatching,
  title={Flow Matching for Generative Modeling},
  author={Lipman, Yaron and Chen, Ricky TQ and Ben-Hamu, Heli and Nickel, Maximilian and Le, Matthew},
  booktitle={The Eleventh International Conference on Learning Representations},
  year={2023}
}

@inproceedings{dit,
  title={Scalable diffusion models with transformers},
  author={Peebles, William and Xie, Saining},
  booktitle={Proceedings of the IEEE/CVF international conference on computer vision},
  pages={4195--4205},
  year={2023}
}

@inproceedings{irasim,
  title={Irasim: A fine-grained world model for robot manipulation},
  author={Zhu, Fangqi and Wu, Hongtao and Guo, Song and Liu, Yuxiao and Cheang, Chilam and Kong, Tao},
  booktitle={Proceedings of the IEEE/CVF International Conference on Computer Vision},
  pages={9834--9844},
  year={2025}
}

@inproceedings{genie,
  title={Genie: Generative interactive environments},
  author={Bruce, Jake and Dennis, Michael D and Edwards, Ashley and Parker-Holder, Jack and Shi, Yuge and Hughes, Edward and Lai, Matthew and Mavalankar, Aditi and Steigerwald, Richie and Apps, Chris and others},
  booktitle={Forty-first International Conference on Machine Learning},
  year={2024}
}

@article{cosmos,
  title={Cosmos world foundation model platform for physical ai},
  author={Agarwal, Niket and Ali, Arslan and Bala, Maciej and Balaji, Yogesh and Barker, Erik and Cai, Tiffany and Chattopadhyay, Prithvijit and Chen, Yongxin and Cui, Yin and Ding, Yifan and others},
  journal={arXiv preprint arXiv:2501.03575},
  year={2025}
}

@article{gigaworld,
  title={Gigaworld-0: World models as data engine to empower embodied ai},
  author={Team, GigaWorld and Ye, Angen and Wang, Boyuan and Ni, Chaojun and Huang, Guan and Zhao, Guosheng and Li, Haoyun and Zhu, Jiagang and Li, Kerui and Xu, Mengyuan and others},
  journal={arXiv preprint arXiv:2511.19861},
  year={2025}
}

@article{emu35,
  title={Emu3.5: Native multimodal models are world learners},
  author={Cui, Yufeng and Chen, Honghao and Deng, Haoge and Huang, Xu and Li, Xinghang and Liu, Jirong and Liu, Yang and Luo, Zhuoyan and Wang, Jinsheng and Wang, Wenxuan and others},
  journal={arXiv preprint arXiv:2510.26583},
  year={2025}
}

@article{dreamerv3,
  title={Mastering diverse control tasks through world models},
  author={Hafner, Danijar and Pasukonis, Jurgis and Ba, Jimmy and Lillicrap, Timothy},
  journal={Nature},
  volume={640},
  number={8059},
  pages={647--653},
  year={2025},
}

@article{mimicgen,
  title={Mimicgen: A data generation system for scalable robot learning using human demonstrations},
  author={Mandlekar, Ajay and Nasiriany, Soroush and Wen, Bowen and Akinola, Iretiayo and Narang, Yashraj and Fan, Linxi and Zhu, Yuke and Fox, Dieter},
  journal={arXiv preprint arXiv:2310.17596},
  year={2023}
}

@article{interna1,
  title={Interndata-a1: Pioneering high-fidelity synthetic data for pre-training generalist policy},
  author={Tian, Yang and Yang, Yuyin and Xie, Yiman and Cai, Zetao and Shi, Xu and Gao, Ning and Liu, Hangxu and Jiang, Xuekun and Qiu, Zherui and Yuan, Feng and others},
  journal={arXiv preprint arXiv:2511.16651},
  year={2025}
}

@article{agibotworld,
  title={Agibot world colosseo: A large-scale manipulation platform for scalable and intelligent embodied systems},
  author={Bu, Qingwen and Cai, Jisong and Chen, Li and Cui, Xiuqi and Ding, Yan and Feng, Siyuan and Gao, Shenyuan and He, Xindong and Hu, Xuan and Huang, Xu and others},
  journal={arXiv preprint arXiv:2503.06669},
  year={2025}
}

@article{droid,
  title={Droid: A large-scale in-the-wild robot manipulation dataset},
  author={Khazatsky, Alexander and Pertsch, Karl and Nair, Suraj and Balakrishna, Ashwin and Dasari, Sudeep and Karamcheti, Siddharth and Nasiriany, Soroush and Srirama, Mohan Kumar and Chen, Lawrence Yunliang and Ellis, Kirsty and others},
  journal={arXiv preprint arXiv:2403.12945},
  year={2024}
}

@article{rtc,
  title={Real-time execution of action chunking flow policies},
  author={Black, Kevin and Galliker, Manuel Y and Levine, Sergey},
  journal={arXiv preprint arXiv:2506.07339},
  year={2025}
}

\newpage
\appendix

\section{Detailed Algorithms}
\label{app:algorithms}

We provide the complete algorithmic procedures for \method. Algorithm~\ref{alg:single_step} details the single denoising step, which jointly processes the multi-modal sequence through the shared DiT trunk and the interleaved depth branch to produce velocity predictions and depth estimates. Algorithm~\ref{alg:ans} presents the full Asynchronous Noise Sampling (ANS) procedure for both training and inference, illustrating how coupled noise sampling during training aligns with the asynchronous denoising schedule at inference time.

\begin{algorithm}[h]
\caption{\textsc{Denoise}: Single Denoising Step of \method}
\label{alg:single_step}
\begin{algorithmic}[1]
\Require Noisy video latent $\mathbf{z}_O^{t_O}$, noisy state $\mathbf{z}_s^{t_a}$, noisy action $\mathbf{z}_a^{t_a}$, video timestep $t_O$, action timestep $t_a$, initial observation $O_0$, initial state $s_0$, language instruction $c$
\Ensure Predicted velocities $\hat{\mathbf{v}}_O, \hat{\mathbf{v}}_s, \hat{\mathbf{v}}_a$; predicted inverse depth $\hat{D}$
\State $\mathbf{z}_{O_0} \gets \mathrm{CausalVAE}(O_0)$;\; $\mathbf{z}_{s_0} \gets \mathrm{MLP}_s(s_0)$ \Comment{Encode conditions with $t = 0$}
\State $\mathbf{Z} \gets \mathrm{Concat}(\mathbf{z}_{O_0},\, \mathbf{z}_O^{t_O},\, \mathbf{z}_{s_0},\, \mathbf{z}_s^{t_a},\, \mathbf{z}_a^{t_a})$
\State Add learnable view embeddings to $\mathbf{Z}$
\For{$i = 1$ \textbf{to} $N - M$} \Comment{Shared trunk}
    \State $\mathbf{Z} \gets \mathrm{DiTBlock}_i(\mathbf{Z})$
\EndFor
\State $\mathbf{Z}_{\text{m}} \gets \mathbf{Z}$;\; $\mathbf{Z}_D \gets \mathbf{Z}$ \Comment{Initialize main and depth branches}
\For{$j = 1$ \textbf{to} $M$} \Comment{Interleaved main-depth processing}
    \State $\mathbf{Z}_D \gets \mathrm{DepthBlock}_j(\mathbf{Z}_D \mid \mathbf{Z}_{\text{m}})$ \Comment{Depth attends to main branch's input}
    \State $\mathbf{Z}_{\text{m}} \gets \mathrm{DiTBlock}_{N-M+j}(\mathbf{Z}_{\text{m}})$ \Comment{Main branch}
\EndFor
\State $\hat{\mathbf{v}}_O,\, \hat{\mathbf{v}}_s,\, \hat{\mathbf{v}}_a \gets \mathrm{Head}_{\text{main}}(\mathbf{Z}_{\text{m}})$
\State $\hat{D} \gets \mathrm{Head}_{\text{depth}}(\mathbf{Z}_D)$ \Comment{Regress inverse depth}
\State \Return $\hat{\mathbf{v}}_O,\, \hat{\mathbf{v}}_s,\, \hat{\mathbf{v}}_a,\, \hat{D}$
\end{algorithmic}
\end{algorithm}

\begin{algorithm}[h]
\caption{Asynchronous Noise Sampling (ANS): Training and Inference}
\label{alg:ans}
\begin{algorithmic}[1]
\Statex \textbf{--- Training: Coupled Noise Sampling ---}
\Require Clean video latent $\mathbf{z}_O^0$, clean state $\mathbf{z}_s^0$, clean action $\mathbf{z}_a^0$, probability $p$
\Ensure Noisy samples $\mathbf{z}_O^{t_O}, \mathbf{z}_s^{t_a}, \mathbf{z}_a^{t_a}$ with coupled timesteps $t_O, t_a$
\State Draw $u \sim \mathrm{U}(0, 1)$
\If{$u < p$} \Comment{Action-conditioned video generation}
    \State $t_a \gets 0$;\; $t_O \sim \mathrm{U}(0, 1)$
\Else \Comment{Asynchronous joint generation}
    \State $t_a \sim \mathrm{U}(0, 1)$;\; $b \sim \mathrm{Beta}(1.5,\, 1)$
    \State $t_O \gets t_a + (1 - t_a) \cdot b$ \Comment{Rescale to $[t_a, 1]$, ensuring $t_O \geq t_a$}
\EndIf
\State $\boldsymbol{\epsilon}_O, \boldsymbol{\epsilon}_s, \boldsymbol{\epsilon}_a \sim \mathcal{N}(\mathbf{0}, \mathbf{I})$
\State $\mathbf{z}_O^{t_O} \gets (1 - t_O)\,\mathbf{z}_O^0 + t_O\,\boldsymbol{\epsilon}_O$;\; $\mathbf{z}_s^{t_a} \gets (1 - t_a)\,\mathbf{z}_s^0 + t_a\,\boldsymbol{\epsilon}_s$;\; $\mathbf{z}_a^{t_a} \gets (1 - t_a)\,\mathbf{z}_a^0 + t_a\,\boldsymbol{\epsilon}_a$ \Comment{Flow matching interpolation}
\State Compute $\mathcal{L}_{\text{total}}$ via \textsc{Denoise} (Algorithm~\ref{alg:single_step}) and backpropagate
\Statex
\Statex \textbf{--- Inference: Asynchronous Denoising ---}
\Require Conditions $(O_0, s_0, c)$, video steps $T_O$, action steps $T_a$ ($T_a < T_O$)
\Ensure Denoised video $\mathbf{z}_O$, state $\mathbf{z}_s$, action $\mathbf{z}_a$
\State $\mathbf{z}_O, \mathbf{z}_s, \mathbf{z}_a \sim \mathcal{N}(\mathbf{0}, \mathbf{I})$ \Comment{Initialize from pure noise}
\State Initialize schedulers: $\mathcal{S}_O$ with $T_O$ steps, $\mathcal{S}_a$ with $T_a$ steps
\For{$k = 1$ \textbf{to} $T_O$}
    \State Get current timestep $t_O$ from $\mathcal{S}_O$
    \If{$k \leq T_a$} \Comment{Joint denoising phase}
        \State Get current timestep $t_a$ from $\mathcal{S}_a$
        \State $\hat{\mathbf{v}}_O,\, \hat{\mathbf{v}}_s,\, \hat{\mathbf{v}}_a,\, \_ \gets \textsc{Denoise}(\mathbf{z}_O, \mathbf{z}_s, \mathbf{z}_a, t_O, t_a, O_0, s_0, c)$
        \State $\mathbf{z}_O \gets \mathcal{S}_O\text{.step}(\mathbf{z}_O, \hat{\mathbf{v}}_O)$;\; $\mathbf{z}_s \gets \mathcal{S}_a\text{.step}(\mathbf{z}_s, \hat{\mathbf{v}}_s)$;\; $\mathbf{z}_a \gets \mathcal{S}_a\text{.step}(\mathbf{z}_a, \hat{\mathbf{v}}_a)$
    \Else \Comment{Video-only denoising phase (action-conditioned)}
        \State $\hat{\mathbf{v}}_O,\, \_,\, \_,\, \_ \gets \textsc{Denoise}(\mathbf{z}_O, \mathbf{z}_s, \mathbf{z}_a, t_O, 0, O_0, s_0, c)$
        \State $\mathbf{z}_O \gets \mathcal{S}_O\text{.step}(\mathbf{z}_O, \hat{\mathbf{v}}_O)$
    \EndIf
\EndFor
\State \Return $\mathbf{z}_O,\, \mathbf{z}_s,\, \mathbf{z}_a$ \Comment{Actions available after step $T_a$; video after step $T_O$}
\end{algorithmic}
\end{algorithm}

\section{Training Details}
\label{app:training_details}

\subsection{Pretraining Data}

Table~\ref{tab:pretrain_data} summarizes the pretraining datasets used in \method. The data spans both real-robot and simulated environments, totaling over 1.49 million episodes and approximately 5{,}874 hours. All datasets undergo careful preprocessing and filtering: we remove episodes containing base locomotion, dexterous manipulation, and failed executions. Following~\cite{pi0}, we additionally filter out stationary frames from the DROID dataset. All videos are uniformly downsampled to 3.75 FPS and resized to a resolution of $320 \times 256$. Since most pretraining datasets lack depth annotations, we extract depth maps from all training videos using Video Depth Anything~\cite{vda}.

\begin{table}[h]
\centering
\caption{Summary of pretraining datasets used in \method.}
\label{tab:pretrain_data}
\small
\begin{tabular}{@{}lcrr@{}}
\toprule
Dataset & Source & Episodes & Duration (h) \\
\midrule
AgibotWorld-Beta~\cite{agibotworld} & Real & 866{,}562 & 2{,}221.5 \\
DROID~\cite{droid} & Real & 74{,}734 & 280.3 \\
InternA1-Aloha~\cite{interna1} & Sim & 184{,}803 & 1{,}337.3 \\
InternA1-Genie1~\cite{interna1} & Sim & 50{,}638 & 174.0 \\
InternA1-Lift2~\cite{interna1} & Sim & 231{,}018 & 1{,}464.7 \\
RoboCasa MimicGen~\cite{robocasa, mimicgen} & Sim & 56{,}771 & 282.4 \\
RoboTwin 2.0~\cite{robotwin2} & Sim & 27{,}500 & 113.7 \\
\midrule
Total & -- & 1{,}492{,}026 & 5{,}873.9 \\
\bottomrule
\end{tabular}
\end{table}

\subsection{Implementation Details}

\paragraph{State and action representation.} To unify heterogeneous single-arm and dual-arm robots across datasets, we define a universal state and action interface based on end-effector poses. The state is represented as a 16-dimensional absolute vector: (position$_3$ + quaternion$_4$ + gripper$_1$) $\times$ 2 arms. The action is represented as a 14-dimensional relative vector: ($\Delta$position$_3$ + $\Delta$axis-angle$_3$ + gripper action$_1$) $\times$ 2 arms. For single-arm robots, only the first 8 dimensions of the state and the first 7 dimensions of the action are supervised. We compute per-dataset quantile statistics ($q_{0.01}$, $q_{0.99}$) for normalization. Notably, the action normalization applies only scaling without bias, preserving the semantics that a zero action corresponds to no movement across all datasets.

\paragraph{Large-scale pretraining.} We pretrain \method on 256 NVIDIA H20 GPUs with a per-GPU batch size of 8 (total batch size 2{,}048). We use the AdamW optimizer with a peak learning rate of $1 \times 10^{-4}$, 1{,}000 steps of linear warmup followed by cosine decay to 0, and train for 40{,}000 steps. The prediction horizon is set to $H = $ \texttt{8}. The loss weighting coefficients are $\lambda_s = $ \texttt{1.0}, $\lambda_a = $ \texttt{1.0}, and $\lambda_D = $ \texttt{1.0}. The number of replicated depth blocks is $M = $ \texttt{10}, and the ANS action-conditioned probability is $p = $ \texttt{0.5}.

\paragraph{Benchmark fine-tuning.} For RoboCasa and RoboTwin 2.0, we further fine-tune the pretrained model on the respective benchmark data using 32 NVIDIA H20 GPUs with a per-GPU batch size of 4 (total batch size 128), a learning rate of $3 \times 10^{-5}$, and the same warmup and cosine decay schedule. Fine-tuning proceeds for 20{,}000 steps. To obtain ground-truth depth maps for fine-tuning, we replay the official demonstration data in the simulator, ensuring that the total data volume and initial configurations remain unchanged and that the replay random seeds do not overlap with those used at test time. For RoboCasa, we directly use the raw actions provided in the dataset as supervision signals. For RoboTwin 2.0, we use relative actions that are converted to absolute end-effector poses based on the state of the first frame in each action chunk before being sent to the simulator for execution.

\paragraph{Inference.} We use asynchronous denoising with $T_a = 10$ action denoising steps and $T_O = 50$ video denoising steps, following the UniPC~\cite{unipc} scheduler. The classifier-free guidance scale is set to 1.0, as we empirically find that larger guidance scales do not improve action quality but increase inference cost. For both benchmarks, each task is evaluated over 100 episodes and the success rate is averaged. All other evaluation settings follow the official benchmark protocols.

\subsection{Baseline Details}

\paragraph{RoboCasa.} The results of $\pi_0$~\cite{pi0}, GR00T-N1.5~\cite{gr00t}, UWM~\cite{uwm}, and Cosmos Policy~\cite{cosmospolicy} are directly taken from~\cite{cosmospolicy}. DreamZero~\cite{dreamzero} is reproduced using the official codebase, with the backbone replaced by Wan2.2-5B~\cite{wan} for fair comparison.

\paragraph{RoboTwin 2.0.} The results of $\pi_0$~\cite{pi0} and $\pi_{0.5}$~\cite{pi0.5} are taken from~\cite{lingbotva}. The results of Motus~\cite{motus} and GigaWorld-Policy~\cite{gigaworldpolicy} are taken from their respective papers. We reimplement UWM~\cite{uwm} with the backbone replaced by Wan2.2-5B~\cite{wan} for fair comparison.

\paragraph{Pretraining data parity.} We note that among all baselines, only DreamZero is fine-tuned from a general-purpose video generation model without prior exposure to robotic data. All other baselines ($\pi_0$, $\pi_{0.5}$, GR00T-N1.5, UWM, Cosmos Policy, GigaWorld-Policy, and Motus) incorporate large-scale robotic datasets in their pretraining pipelines. Therefore, the comparison is conducted under broadly comparable data regimes.

\section{Detailed Results}
\label{app:detailed_results}

We report per-task success rates for \method on the RoboCasa and RoboTwin 2.0 benchmarks.

\subsection{Per-Task Results on RoboCasa}

Table~\ref{tab:robocasa_detailed} presents the success rate of \method on each of the 24 manipulation tasks in the RoboCasa benchmark.

\begin{table}[h]
\centering
\caption{Per-task success rate (\%) of \method on the RoboCasa benchmark (24 tasks).}
\label{tab:robocasa_detailed}
\small
\begin{tabular}{@{}clc@{}}
\toprule
\# & Task & SR \\
\midrule
1 & \texttt{CloseDoubleDoor} & 87.0 \\
2 & \texttt{CloseDrawer} & 100.0 \\
3 & \texttt{CloseSingleDoor} & 96.0 \\
4 & \texttt{CoffeePressButton} & 96.0 \\
5 & \texttt{CoffeeServeMug} & 82.0 \\
6 & \texttt{CoffeeSetupMug} & 45.0 \\
7 & \texttt{OpenDoubleDoor} & 94.0 \\
8 & \texttt{OpenDrawer} & 85.0 \\
9 & \texttt{OpenSingleDoor} & 96.0 \\
10 & \texttt{PnPCabToCounter} & 73.0 \\
11 & \texttt{PnPCounterToCab} & 67.0 \\
12 & \texttt{PnPCounterToMicrowave} & 62.0 \\
13 & \texttt{PnPCounterToSink} & 79.0 \\
14 & \texttt{PnPCounterToStove} & 83.0 \\
15 & \texttt{PnPMicrowaveToCounter} & 57.0 \\
16 & \texttt{PnPSinkToCounter} & 71.0 \\
17 & \texttt{PnPStoveToCounter} & 80.0 \\
18 & \texttt{TurnOffMicrowave} & 93.0 \\
19 & \texttt{TurnOffSinkFaucet} & 86.0 \\
20 & \texttt{TurnOffStove} & 35.0 \\
21 & \texttt{TurnOnMicrowave} & 82.0 \\
22 & \texttt{TurnOnSinkFaucet} & 92.0 \\
23 & \texttt{TurnOnStove} & 80.0 \\
24 & \texttt{TurnSinkSpout} & 80.0 \\
\midrule
& Average & 79.2 \\
\bottomrule
\end{tabular}
\end{table}

\subsection{Per-Task Results on RoboTwin 2.0}

Table~\ref{tab:robotwin_detailed} presents the per-task success rate under both Clean and Randomized settings on RoboTwin 2.0.

\begin{table}[h]
\centering
\caption{Per-task success rate (\%) of \method on the RoboTwin 2.0 benchmark (50 tasks).}
\label{tab:robotwin_detailed}
\footnotesize
\begin{minipage}[t]{0.48\textwidth}
\centering
\begin{tabular}{@{}clcc@{}}
\toprule
\# & Task & Clean & Rand. \\
\midrule
1 & \texttt{adjust\_bottle} & 100.0 & 99.0 \\
2 & \texttt{beat\_block\_hammer} & 98.0 & 96.0 \\
3 & \texttt{blocks\_ranking\_rgb} & 99.0 & 95.0 \\
4 & \texttt{blocks\_ranking\_size} & 76.0 & 82.0 \\
5 & \texttt{click\_alarmclock} & 98.0 & 99.0 \\
6 & \texttt{click\_bell} & 100.0 & 100.0 \\
7 & \texttt{dump\_bin\_bigbin} & 90.0 & 96.0 \\
8 & \texttt{grab\_roller} & 100.0 & 100.0 \\
9 & \texttt{handover\_block} & 88.0 & 79.0 \\
10 & \texttt{handover\_mic} & 88.0 & 89.0 \\
11 & \texttt{hanging\_mug} & 46.0 & 55.0 \\
12 & \texttt{lift\_pot} & 100.0 & 99.0 \\
13 & \texttt{move\_can\_pot} & 80.0 & 84.0 \\
14 & \texttt{move\_pillbottle\_pad} & 96.0 & 98.0 \\
15 & \texttt{move\_playingcard\_away} & 100.0 & 98.0 \\
16 & \texttt{move\_stapler\_pad} & 67.0 & 70.0 \\
17 & \texttt{open\_laptop} & 96.0 & 97.0 \\
18 & \texttt{open\_microwave} & 89.0 & 92.0 \\
19 & \texttt{pick\_diverse\_bottles} & 91.0 & 92.0 \\
20 & \texttt{pick\_dual\_bottles} & 99.0 & 100.0 \\
21 & \texttt{place\_a2b\_left} & 90.0 & 87.0 \\
22 & \texttt{place\_a2b\_right} & 92.0 & 89.0 \\
23 & \texttt{place\_bread\_basket} & 90.0 & 91.0 \\
24 & \texttt{place\_bread\_skillet} & 90.0 & 96.0 \\
25 & \texttt{place\_burger\_fries} & 97.0 & 99.0 \\
\bottomrule
\end{tabular}
\end{minipage}
\hfill
\begin{minipage}[t]{0.48\textwidth}
\centering
\begin{tabular}{@{}clcc@{}}
\toprule
\# & Task & Clean & Rand. \\
\midrule
26 & \texttt{place\_can\_basket} & 84.0 & 82.0 \\
27 & \texttt{place\_cans\_plasticbox} & 99.0 & 98.0 \\
28 & \texttt{place\_container\_plate} & 98.0 & 100.0 \\
29 & \texttt{place\_dual\_shoes} & 83.0 & 81.0 \\
30 & \texttt{place\_empty\_cup} & 98.0 & 99.0 \\
31 & \texttt{place\_fan} & 84.0 & 92.0 \\
32 & \texttt{place\_mouse\_pad} & 84.0 & 86.0 \\
33 & \texttt{place\_object\_basket} & 85.0 & 87.0 \\
34 & \texttt{place\_object\_scale} & 93.0 & 89.0 \\
35 & \texttt{place\_object\_stand} & 97.0 & 96.0 \\
36 & \texttt{place\_phone\_stand} & 75.0 & 80.0 \\
37 & \texttt{place\_shoe} & 97.0 & 99.0 \\
38 & \texttt{press\_stapler} & 94.0 & 90.0 \\
39 & \texttt{put\_bottles\_dustbin} & 85.0 & 95.0 \\
40 & \texttt{put\_object\_cabinet} & 66.0 & 76.0 \\
41 & \texttt{rotate\_qrcode} & 84.0 & 83.0 \\
42 & \texttt{scan\_object} & 86.0 & 79.0 \\
43 & \texttt{shake\_bottle} & 99.0 & 99.0 \\
44 & \texttt{shake\_bottle\_horiz.} & 100.0 & 99.0 \\
45 & \texttt{stack\_blocks\_three} & 97.0 & 95.0 \\
46 & \texttt{stack\_blocks\_two} & 100.0 & 100.0 \\
47 & \texttt{stack\_bowls\_three} & 88.0 & 82.0 \\
48 & \texttt{stack\_bowls\_two} & 98.0 & 98.0 \\
49 & \texttt{stamp\_seal} & 93.0 & 95.0 \\
50 & \texttt{turn\_switch} & 61.0 & 72.0 \\
\midrule
& Average & 89.8 & 90.7 \\
\bottomrule
\end{tabular}
\end{minipage}
\end{table}

\section{Real Robot Experiments}
\label{app:real_robot}

To validate the practical applicability of \method, we deploy the model on a real-world dual-arm robotic platform and evaluate it on an earphone packing task. We select this task because it is a challenging long-horizon manipulation scenario that places stringent demands on 3D spatial reasoning: the robot must estimate the precise 6-DoF pose of the earphone case, localize the narrow slot openings whose geometry varies across orientations, and execute insertion trajectories that satisfy sub-centimeter clearance constraints in all three spatial dimensions. These requirements make earphone packing a compelling testbed for evaluating whether the explicit 4D spatial awareness provided by \method translates into tangible manipulation benefits compared to methods that operate purely in 2D pixel space. Beyond spatial precision, the task further requires robust bimanual coordination across four sequential stages, testing both long-horizon reliability and real-time execution efficiency.

\begin{figure}[h]
\centering
\includegraphics[width=\textwidth]{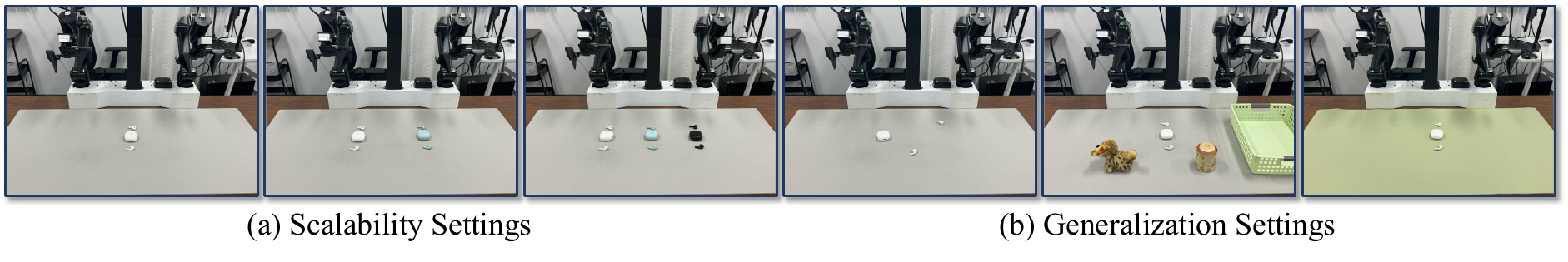}
\vspace{-0.2in}
\caption{Real-world experimental setup. The AC One dual-arm platform is equipped with one main camera providing a global view and two wrist-mounted cameras for close-up observations. The earphone packing task requires precise bimanual coordination and robust insertion under tight geometric tolerances.}
\label{fig:real_setup}
\end{figure}

\paragraph{Setup.} All experiments are conducted on an AC One dual-arm platform equipped with one main camera and two wrist-mounted cameras, all operating at a resolution of $320 \times 256$, as illustrated in Figure~\ref{fig:real_setup}. We collect approximately 20 hours of demonstration data for the earphone packing task. The model is fine-tuned on 64 NVIDIA H20 GPUs with a per-GPU batch size of 4 for 40{,}000 steps. We employ asynchronous inference with 8 denoising steps on an NVIDIA RTX 5090 D GPU, yielding a single-pass latency of approximately 300\,ms per action chunk. We further adopt the Real-Time Chunking (RTC) method~\cite{rtc} to overlap denoising computation with action execution. The robot operates at a control frequency of 15\,Hz, executing 15 actions (1 second) per chunk with an RTC inference delay of 6 actions, enabling seamless real-time deployment.

\paragraph{Task design.} Since earphone packing is a multi-step long-horizon task, we decompose it into four sequential stages, each contributing 25\% of the total progress:
\begin{enumerate}
    \item Grasp the empty earphone case and open the lid.
    \item Pick up one earphone and correctly place it into the case.
    \item Pick up the other earphone and correctly place it into the case.
    \item Close the lid and return the case to the table.
\end{enumerate}
Successfully completing all four stages constitutes 100\% progress. We evaluate under six settings designed to test both scalability and generalization:
\begin{itemize}
    \item \textbf{Scalability}: consecutively packing 1, 2, or 3 earphones in a single episode, testing the model's ability to handle increasing task length.
    \item \textbf{Generalization}: packing 1 earphone under three out-of-distribution conditions not seen during training: (i) novel object placements, (ii) unseen tablecloth colors, and (iii) unseen distractor objects.
\end{itemize}
For each setting, we conduct 6 trials: for single-earphone tasks, each of 3 earphone colors is tested twice; for multi-earphone tasks, 6 trials are run with varying color orderings. We report two metrics: average progress (\%) across all episodes, and average completion time (seconds) computed only over episodes that achieve 100\% progress.

\paragraph{Quantitative results.} Table~\ref{tab:real_robot} summarizes the real-robot evaluation results. We compare against Xiaomi-Robotics-0~\cite{xr0}, a recent VLA model featuring strong performance and low inference latency. We also attempted to train $\pi_{0.5}$~\cite{pi0.5} using the openpi\footnote{\url{https://github.com/Physical-Intelligence/openpi}} codebase under the same data and training setup; however, it failed to complete a single full earphone packing episode across all settings, achieving only 25\%--50\% progress on the single-earphone task (corresponding to grasping the case but failing at earphone insertion). We therefore omit $\pi_{0.5}$ from the table and only report XR-0 as the baseline.

\begin{table}[h]
\centering
\caption{Real-robot earphone packing results. Progress (\%) is averaged over all 6 episodes; completion time (s) is averaged over episodes reaching 100\% progress.}
\label{tab:real_robot}
\small
\begin{tabular}{@{}llcccc@{}}
\toprule
& \multirow{2}{*}{Setting} & \multicolumn{2}{c}{XR-0~\cite{xr0}} & \multicolumn{2}{c}{\method (Ours)} \\
\cmidrule(lr){3-4} \cmidrule(lr){5-6}
& & Prog.\,(\%) & Time\,(s) & Prog.\,(\%) & Time\,(s) \\
\midrule
\multicolumn{6}{@{}l}{\textit{Scalability}} \\
& Pack 1 earphone & 100.0 (24/24) & 54.66 & \textbf{100.0 (24/24)} & \textbf{41.63} \\
& Pack 2 earphones & 79.1 (38/48) & 115.44 & \textbf{93.8 (45/48)} & \textbf{113.25} \\
& Pack 3 earphones & 63.9 (46/72) & 195.66 & \textbf{68.0 (49/72)} & \textbf{160.72} \\
\midrule
\multicolumn{6}{@{}l}{\textit{Generalization}} \\
& Novel placements & 58.3 (14/24) & 89.63 & \textbf{70.8 (17/24)} & \textbf{46.68} \\
& Unseen tablecloth & \textbf{66.7 (16/24)} & 65.73 & \textbf{66.7 (16/24)} & \textbf{62.01} \\
& Unseen distractors & 66.7 (16/24) & 76.32 & \textbf{75.0 (18/24)} & \textbf{51.53} \\
\bottomrule
\end{tabular}
\end{table}

\paragraph{Analysis.} As shown in Table~\ref{tab:real_robot}, \method consistently outperforms XR-0 across both scalability and generalization settings. In the scalability evaluation, both methods achieve perfect progress on the single-earphone task, but as the number of earphones increases, \method demonstrates superior long-horizon reliability: achieving 93.8\% progress on packing 2 earphones compared to 79.1\% for XR-0, and 68.0\% vs.\ 63.9\% on packing 3 earphones. We attribute this to the explicit 3D spatial awareness of \method, which provides more accurate geometric reasoning for the precise insertion operations required across successive packing stages. In the generalization evaluation, \method shows stronger robustness under novel placements (70.8\% vs.\ 58.3\%) and unseen distractors (75.0\% vs.\ 66.7\%), while matching XR-0 on unseen tablecloths (66.7\%). The improvement on novel placements is particularly notable, as it requires the model to generalize its spatial reasoning to unseen object configurations. Furthermore, \method consistently achieves lower completion times across all settings, indicating that asynchronous denoising with real-time chunking enables not only faster inference but also more efficient trajectories with fewer corrective motions.

\begin{figure}[h]
\centering
\includegraphics[width=\textwidth]{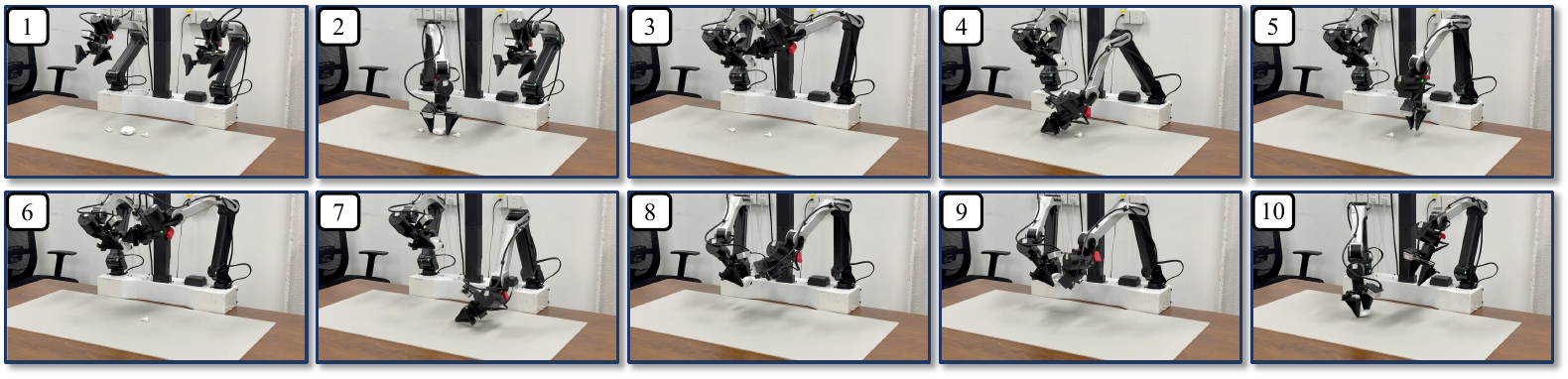}
\caption{Qualitative results of \method deployed on a real AC One dual-arm robot for the earphone packing task. Each image shows keyframes from a representative execution rollout.}
\label{fig:real_robot}
\end{figure}

\paragraph{Qualitative results.} Figure~\ref{fig:real_robot} presents a representative rollout sequence of \method on the earphone packing task. For more qualitative results, please visit our \href{https://sharinka0715.github.io/X-WAM/}{project website}. %supplementary material.

\section{Limitations and Future Work}
\label{app:limitations}

While \method demonstrates strong performance across simulation benchmarks and real-robot deployment, two primary limitations remain.

First, the current framework processes only a fixed-length context window of observations without incorporating historical information or autoregressive rollout, unlike approaches such as DreamZero~\cite{dreamzero} that leverage KV caching for extended temporal context. This limited context horizon may hinder the model's ability to fully comprehend task progress in long-horizon manipulation scenarios, potentially leading to suboptimal decisions when the current observation alone is insufficient to disambiguate the task stage.

Second, as a unified model that jointly generates high-dimensional videos and low-dimensional actions, \method incurs higher inference latency compared to dedicated policy models. Specialized VLAs and lightweight WAMs such as Fast-WAM~\cite{fastwam} achieve substantially lower per-step latency, whereas \method requires approximately 300\,ms per action chunk with 8 denoising steps. Although real-time chunking~\cite{rtc} enables seamless deployment on physical robots by overlapping computation with execution, the additional inference delay can degrade policy performance, as the robot must act on predictions computed several frames in the past.

Both limitations point to promising directions for future work. Our proposed architecture and noise scheduling strategy are orthogonal to long-context mechanisms, and \method can be readily extended with history conditioning, KV caching, or autoregressive inference to support longer temporal horizons. Similarly, advances in inference acceleration, such as model distillation, consistency models, and more aggressive asynchronous scheduling, could further narrow the latency gap with dedicated policy models while preserving the benefits of unified 4D modeling.

% \newpage
% \input{checklist}

\end{document}